\documentclass[sigconf]{acmart}
\AtBeginDocument{%
  \providecommand\BibTeX{{%
    \normalfont B\kern-0.5em{\scshape i\kern-0.25em b}\kern-0.8em\TeX}}}
    
\usepackage[linesnumbered,ruled,vlined]{algorithm2e}
\usepackage{bbold}
\usepackage{mathtools}
\usepackage{multirow}
\usepackage{stfloats}
\usepackage{subfigure}
\usepackage{xspace}

\copyrightyear{2023} 
\acmYear{2023} 
\setcopyright{acmlicensed}\acmConference[WWW '23]{Proceedings of the ACM Web Conference 2023}{April 30-May 4, 2023}{Austin, TX, USA}
\acmBooktitle{Proceedings of the ACM Web Conference 2023 (WWW '23), April 30-May 4, 2023, Austin, TX, USA}
\acmPrice{15.00}
\acmDOI{10.1145/3543507.3583305}
\acmISBN{978-1-4503-9416-1/23/04}

\newcommand{\modelname}{FedLU\xspace}
\DeclareMathOperator*{\argmin}{arg\,min}

\begin{document}

\title{Heterogeneous Federated Knowledge Graph Embedding Learning and Unlearning}

\author{Xiangrong Zhu}
\affiliation{
  \department{State Key Laboratory for Novel Software Technology}
  \institution{Nanjing University \country{China}}
}
\email{xrzhu.nju@gmail.com}

\author{Guangyao Li}
\affiliation{
  \department{State Key Laboratory for Novel Software Technology}
  \institution{Nanjing University \country{China}}
}
\email{gyli.nju@gmail.com}

\author{Wei Hu}
\authornote{Wei Hu is the corresponding author.}
\affiliation{
    \department{State Key Laboratory for Novel Software Technology}
    \department{National Institute of Healthcare\\ Data Science}
    \institution{Nanjing University \country{China}}
}
\email{whu@nju.edu.cn}

\renewcommand{\shortauthors}{Xiangrong Zhu, Guangyao Li, \& Wei Hu}

\begin{abstract}
Federated Learning (FL) recently emerges as a paradigm to train a global machine learning model across distributed clients without sharing raw data.
Knowledge Graph (KG) embedding represents KGs in a continuous vector space, serving as the backbone of many knowledge-driven applications.
As a promising combination, federated KG embedding can fully take advantage of knowledge learned from different clients while preserving the privacy of local data.
However, realistic problems such as data heterogeneity and knowledge forgetting still remain to be concerned.
In this paper, we propose \modelname, a novel FL framework for heterogeneous KG embedding learning and unlearning.
To cope with the drift between local optimization and global convergence caused by data heterogeneity, we propose mutual knowledge distillation to transfer local knowledge to global, and absorb global knowledge back.
Moreover, we present an unlearning method based on cognitive neuroscience, which
combines retroactive interference and passive decay to erase specific knowledge from local clients and propagate to the global model by reusing knowledge distillation.
We construct new datasets for assessing realistic performance of the state-of-the-arts.
Extensive experiments show that \modelname achieves superior results in both link prediction and knowledge forgetting.
\end{abstract}




\maketitle

%==============================
\section{Introduction}
\label{sect:intro}

\emph{Federated Learning} (FL) \cite{federated_machine_learning,FLSurvey} is a new machine learning paradigm, where multiple devices with local data samples work together to train a global machine learning model. 
FL differs significantly from classic centralized machine learning scenarios in that it does not need all local data to be uploaded to a central server. 
It also stands in contrast to traditional distributed methods which often assume that local data samples are identically distributed.
\emph{Knowledge Graph} (KG) is a structured knowledge base that describes real-world entities and their relations in the form of triplets.
Recent advances in representation learning techniques hasten the advent of \emph{KG embedding} \cite{KGSurvey,KGESurvey}, which projects entities and relations into a unified semantic space to mitigate the symbolic heterogeneity and support various knowledge-driven applications.
\emph{Federated KG embedding learning} \cite{FedE,FedEC,FKE,FKGE,FedR} is an emerging field that leverages FL and multi-source KGs for collaborative KG embedding.

\begin{figure}
  \centering
  \includegraphics[width=.94\linewidth]{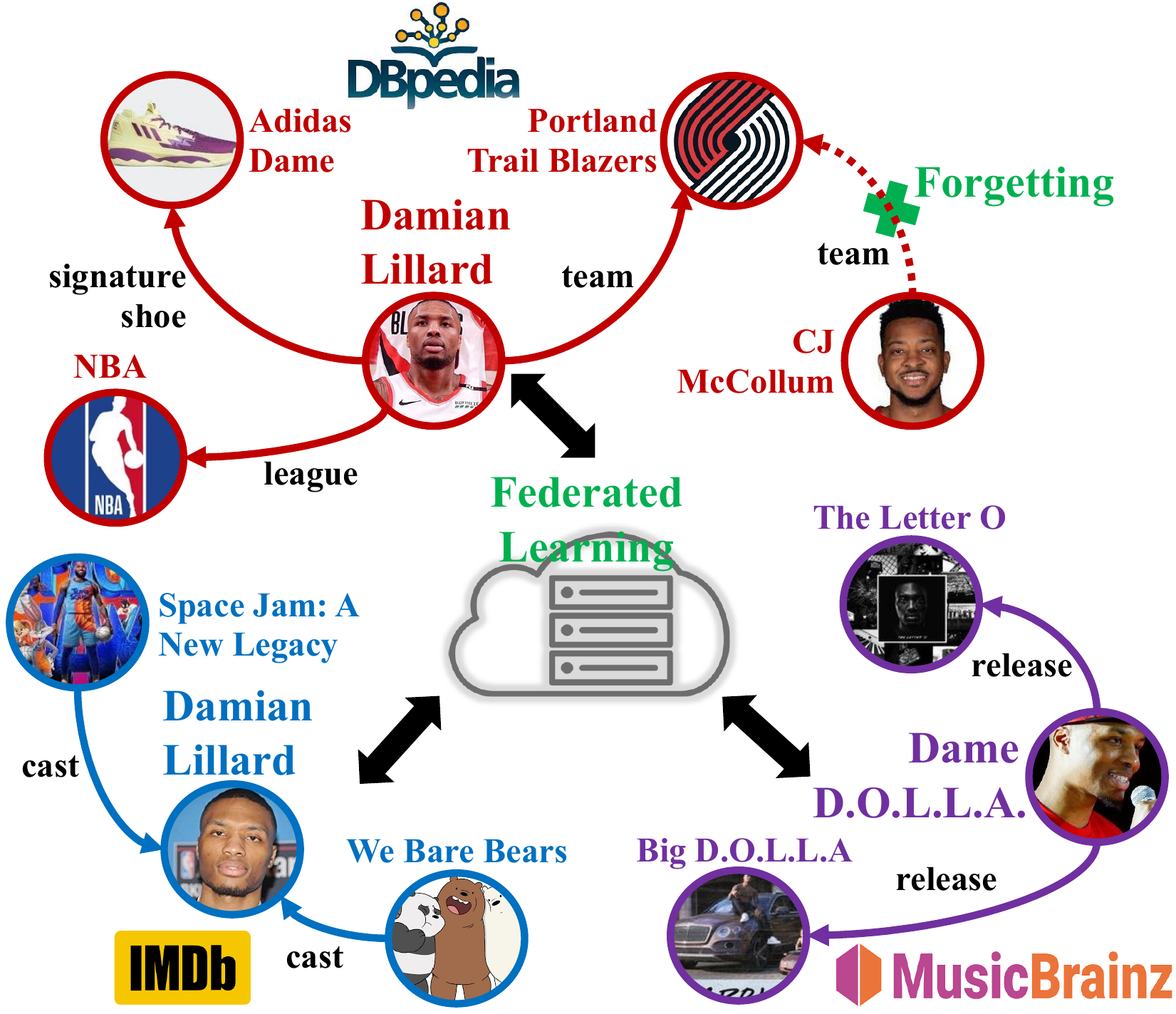}
  \caption{A motivating example, where different KGs (e.g., DBpedia, MusicBrainz and IMDb) provide complementary knowledge about the NBA all-star \textit{Damian Lillard}.}
  \label{fig:motivation}
\end{figure}

\smallskip
\noindent\textsc{Example 1.}
Let us see the example in Figure~\ref{fig:motivation}.
The NBA all-star, \textit{Damian Lillard}, has made great achievements in basketball, sport business, film and music, documented in distributed KGs like DBpedia, IMDb and MusicBrainz.
Based on FL, we can learn the above knowledge jointly and securely to better model the overall image of \textit{Damian Lillard}.
However, federated KG representation learning needs to address realistic challenges like data heterogeneity and knowledge forgetting.

\smallskip
\noindent\textsc{Challenge 1:} \textbf{Data heterogeneity.} A major difference between FL and traditional distributed learning is that the data feature and distribution in FL are unknown and uncontrollable, while traditional distributed learning assumes that local data samples are identically distributed.
KGs combine the characteristics of relational databases and graph data, so their heterogeneity is particularly significant.
On one hand, the schemata of different KGs describe various domains with uneven abstraction levels. 
As illustrated in Figure~\ref{fig:motivation}, the entities and relations instantiating these KGs also do not fully overlap.
On the other hand, the graph features of different KGs such as degree distribution and clustering coefficient are inconsistent.
Local data with strong heterogeneity make local KG embedding models difficult to be aggregated, leading to the drift between local optimization and global convergence, and therefore reduce the overall accuracy of federated KG embedding learning.
For example, as a basketball player, \textit{Damian Lillard}'s status may make him more likely to be associated with the sport drama ``The Way Back'', while suppress the prediction that he is a voice actor in the comedy animation ``We Bare Bears''.

\smallskip
\noindent\textsc{Challenge 2:} \textbf{Knowledge forgetting.} As the symbolic abstraction of real-world domains, local KGs on clients can be outdated or erroneous, leading to the global model change over time.
For example, \textit{Damian Lillard} and \textit{CJ McCollum} are no longer teammates in the Portland Trail Blazers, as \textit{CJ McCollum} was traded to the New Orleans Pelicans.
Another need for knowledge forgetting is privacy protection in healthcare, finance, insurance, etc.
For instance, the EU General Data Protection Regulation (GDPR) guarantees users' right to be forgotten,\footnote{\url{https://gdpr-info.eu/art-17-gdpr/}} i.e., the data subject shall have the right to obtain from the controller the erasure of personal data concerning him or her without undue delay and the controller shall have the obligation to erase personal data without undue delay.
It is easy to remove triplets in a KG, but hard to eliminate their impact on the trained embedding model.
Moreover, re-training the model again on the remaining data is time-consuming and computationally expensive.
The difficulty of making the KG embedding model forget specific triplets lies in that it is hard to quantify and recover the influence of a triplet on the model due to the correlation with other triplets and the randomness of training.
Given that local data are not visible to the global view, how to propagate the unlearning results to the global model is quite challenging.

\smallskip
\noindent\textbf{Current approaches.} 
Existing work mainly focuses on the migration and customization from FL to KG embedding, which can be divided into client-server \cite{FedE,FedR,FKE,FedEC} and peer-to-peer \cite{FKGE} based on the communication architecture of nodes in the whole system.
The client-server architecture typically consists of one server and multiple clients with their own local KGs.
The server initializes the embeddings of entities and relations, distributes the embeddings to clients for training, and aggregates the clients' locally-trained embedding models after each round of local training.
In contrast, the peer-to-peer architecture allows clients to exchange embeddings and learn from each other in a generative adversarial manner.
As far as FL is concerned, the client-server architecture has higher communication efficiency and can eventually produce a common and effective global model for general use.
However, it is difficult for existing federated KG embedding learning methods to deal with non-IID data with strong heterogeneity.
See Section~\ref{subsect:dataset} for an analysis of the datasets used in existing work.
Furthermore, to our best knowledge, we have not seen any exploration in KG embedding unlearning, let alone making the global model forget specific triplets from clients under the FL setting.
% Furthermore, to our best knowledge, we have not seen any exploration in KG embedding unlearning, let alone the propagation from local embeddings to global models under the FL setting.

\smallskip
\noindent\textbf{Our framework.} In this paper, we consider the realistic challenges in federated KG embedding, and propose a novel FL framework for heterogeneous KG embedding learning and unlearning, dubbed \textbf{\modelname}.
To address the data heterogeneity of multi-source KGs, we propose mutual knowledge distillation to transfer local knowledge to global, and absorb global knowledge back.
Furthermore, to achieve knowledge forgetting, we present an unlearning method to erase specific knowledge from local embeddings and propagate to the global embedding by reusing knowledge distillation.
To validate the effectiveness of the proposed framework, we construct three new datasets based on FB15k-237 and carry out extensive experiments with varied number of clients.

The main contributions of this paper are summarized as follows:
\begin{itemize}
    \item We propose a FL framework for KG embedding learning. 
    In particular, we design a mutual knowledge distillation method to cope with the drift between local optimization and global convergence caused by data heterogeneity. (See Section~\ref{subsect:learn})
    
    \item Based on cognitive neuroscience, we present a novel KG embedding unlearning method, which combines retroactive interference and passive decay to achieve knowledge forgetting. (See Section~\ref{subsect:unlearn})
    
    \item We conduct extensive experiments on newly-constructed datasets with varied number clients.
    Experimental results show that \modelname outperforms the state-of-the-arts in both link prediction and knowledge forgetting. (See Section~\ref{sect:exp})
\end{itemize}

%==============================
\section{Preliminaries}
\label{sect:prelim}

%In this section, we present necessary preliminaries to federated KG embedding learning and unlearning.

%--------------------
\noindent\textbf{Federated learning.}
FL is a computing paradigm where disperse clients collaboratively train models using multi-party datasets without data leakage.
Instead of centrally collecting data into a machine learning server or sharing training data across parties, FL only involves an exchange of model parameters or gradients. 
FedAvg~\cite{FedAvg} is a very popular FL algorithm due to its simplicity and effectiveness, where each client trains its local model with private data, and the server aggregates the uploaded local models by a weighted sum.
The procedure can be denoted by $\mathbf{w}_{t+1} = \sum_{k\in K} \frac{n_k}{n}\mathbf{w}^k_t$, where $\mathbf{w}_t$ denotes the aggregated global model at the $t$-th iteration, and $\mathbf{w}_t^k$ denotes the optimized local model for client $k$. 
$\frac{n_k}{n}$ is the aggregation weight considering the proportion of the dataset size for each client.
It is proved that FedAvg can converge on non-IID data under necessary conditions \cite{convergence_of_FedAvg}.
In the real world, the system and statistical heterogeneity widely exists, such as device failures and data biases.
In order to alleviate such heterogeneity, FedProx \cite{FedProx} introduces a proximal term to the local training process as a minor modification to FedAvg.
The local update can be denoted by $\mathbf{w}^k_{t+1} = \argmin_\mathbf{w} F_k(\mathbf{w}) + \frac{\mu}{2}||\,\mathbf{w}-\mathbf{w}_t\,||^2$,
where $F_k(\cdot)$ is the original local objective function, and $\mu$ is a weight to balance the two terms.
MOON \cite{MOON} presents a model-level contrastive learning to solve the heterogeneity in images with deep learning models. It adds a model-contrastive loss into local optimization to decrease the distance between the local and global models.
In addition, researches increasingly turn to knowledge distillation~\cite{FedMD,FD,FedKD} instead of direct model substitution.
We refer readers to a recent survey \cite{FLSurvey} for more progress.

%--------------------
\smallskip
\noindent\textbf{KG embedding.}
KG embedding models embed entities and relations into a continuous vector space while expecting to reveal the inherent semantics of KGs.
Existing work can be categorized into three groups:
(i) Translational distance models \cite{TransE,RotatE,HAKE} view a relation as a translation operation from the head entity to the tail entity, and exploit a distance-based scoring function to measure the plausibility of a triplet.
TransE \cite{TransE} adopts vector translation to model such an operation.
Its scoring function is defined as $\mathcal{S}_{\text{TransE}}(h,r,t)=-||\,\mathbf{h}+\mathbf{r}-\mathbf{t}\,||$, while RotatE \cite{RotatE} interprets a relation as a vector rotation.
(ii) Semantic matching models \cite{DistMult,ComplEx,TuckER} match latent semantics of entities and relations in their scoring functions.
ComplEx  \cite{ComplEx} captures pairwise interactions between entities and relations along the same dimension in a complex vector space.
The scoring function is defined as $\mathcal{S}_{\text{ComplEx}}(h,r,t)=\text{Re}\big(\langle\mathbf{h} ,\mathbf{r},\mathbf{\bar{t}}\rangle\big)$, where $\text{Re}(\cdot)$ denotes the real vector component. 
$\langle\cdot\rangle$ denotes a generalized dot product. 
$\bar{\cdot}$ is the conjugate of a complex vector.
(iii) Deep models leverage deep neural networks to dig into the underlying structures of KGs, such as convolutional neural networks \cite{ConvE,ConvKB} and graph neural networks \cite{RGCN,CompGCN}.

%--------------------
\smallskip
\noindent\textbf{Machine unlearning.}
Different from machine learning, machine unlearning aims to make models forget specific samples.
The work in \cite{forget_machine_unlearning} transforms the original machine learning algorithm into a statistical query learning form.
Therefore, unlearning can be realized by updating the summations for queries instead of re-training from scratch.
The work in \cite{make_AI_forget_you} further leverages k-means clustering to make the unlearning algorithm in \cite{forget_machine_unlearning} more deletion-efficient. 
Inspired by ensemble learning and incremental learning, SISA \cite{machine_unlearning} first splits the whole dataset into shards to train different submodels, and then splits each shard into slices to feed into the individual submodels sequentially while storing checkpoints. 
%Through the isolation of training data and process, unlearning of a specific data sample only needs to locate the individual submodel and re-train from the last checkpoint not involving it.
To the best of our knowledge, there is no prior work exploring KG embedding unlearning.

%==============================
\section{Related Work}
\label{sect:related}

There is only a few studies on federated KG embedding learning.
To protect data privacy and sensitivity, FedE \cite{FedE} simply adapts FedAvg to KG embedding by sharing entity embeddings. 
The adaptation of delivery and aggregation is defined as follows:
\begin{align}
%\label{eq:fededeliver}
    \mathbf{E}_t^k = \mathbf{P}^{k}\mathbf{E}_t,\quad
%\label{eq:fedeaggregate}
    \mathbf{E}_{t+1} = \Big(\mathbb{1}\oslash \sum_{k=1}^{K}\mathbf{v}^k\Big)\otimes \sum_{k=1}^{K}{\mathbf{P}^k}^\top\mathbf{E}_{t}^{k},
\end{align}
where $\mathbf{E}_t$ is the global embedding at the $t$-th communication round,
$\mathbf{P}^k$ is the permutation matrix from the global embedding to the local embedding in client $k$ and ${\mathbf{P}^k}^\top$ is its transpose,
$\mathbf{E}^k_t$ is the local embedding of client $k$ at this communication round,
$\mathbf{v}^k$ is an existence vector indicating which entities exist in client $k$,
$\mathbb{1}$ refers to an all-one vector,
and $\oslash,\otimes$ are respectively element-wise division for vectors and element-wise multiplication with broadcasting.
$\mathbf{P}^k$ and $\mathbf{v}^k$ can be calculated by the server without sharing raw data through techniques such as private data matching~\cite{PrivateSchemaMatching}.
% FedE empirically validates that the global embeddings can be aggregated into locally-computed embeddings to achieve better performance.
FedE empirically validates that the locally-computed embeddings can achieve better performance by aggregating with the global embedding.

FedR \cite{FedR} pays attention to the scenario where clients have high overlapping not only in entities but also in relations. 
The clients in its framework share relation embeddings globally and represent entity embeddings based on relations locally.
In this paper, we do not assume relation overlapping.

FKGE \cite{FKGE} proposes a peer-to-peer federated embedding learning framework for KGs in various domains, based on a differentially-privacy generative adversarial model.
Borrowing the idea of MUSE \cite{MUSE} to align two manifolds with a translational mapping matrix, FKGE is able to generate refined embeddings after generative adversarial training.
For communication efficiency, we do not adopt the peer-to-peer architecture.

Inspired by MOON~\cite{MOON}, FedEC~\cite{FedEC} uses embedding-contrastive learning to guide the embedding optimization for addressing data heterogeneity. 
%It increases the similarity between local and global embeddings by incorporating a contrastive loss $\mathcal{L}^\text{con}$ into the local loss $\mathcal{L}^\text{predict}$ to formulate $\mathcal{L}=\mathcal{L}^\text{predict}+\mu_{\text{con}}\mathcal{L}^\text{con}$, where $\mu_{\text{con}}$ is a hyper-parameter to control the weight of the contrastive loss.
It increases the similarity between local and global embeddings by incorporating a contrastive loss into the local loss.

%==============================
\section{The Proposed Framework}
\label{sect:framework}

\begin{figure*}[!t]
  \centering
  \includegraphics[width=.86\linewidth]{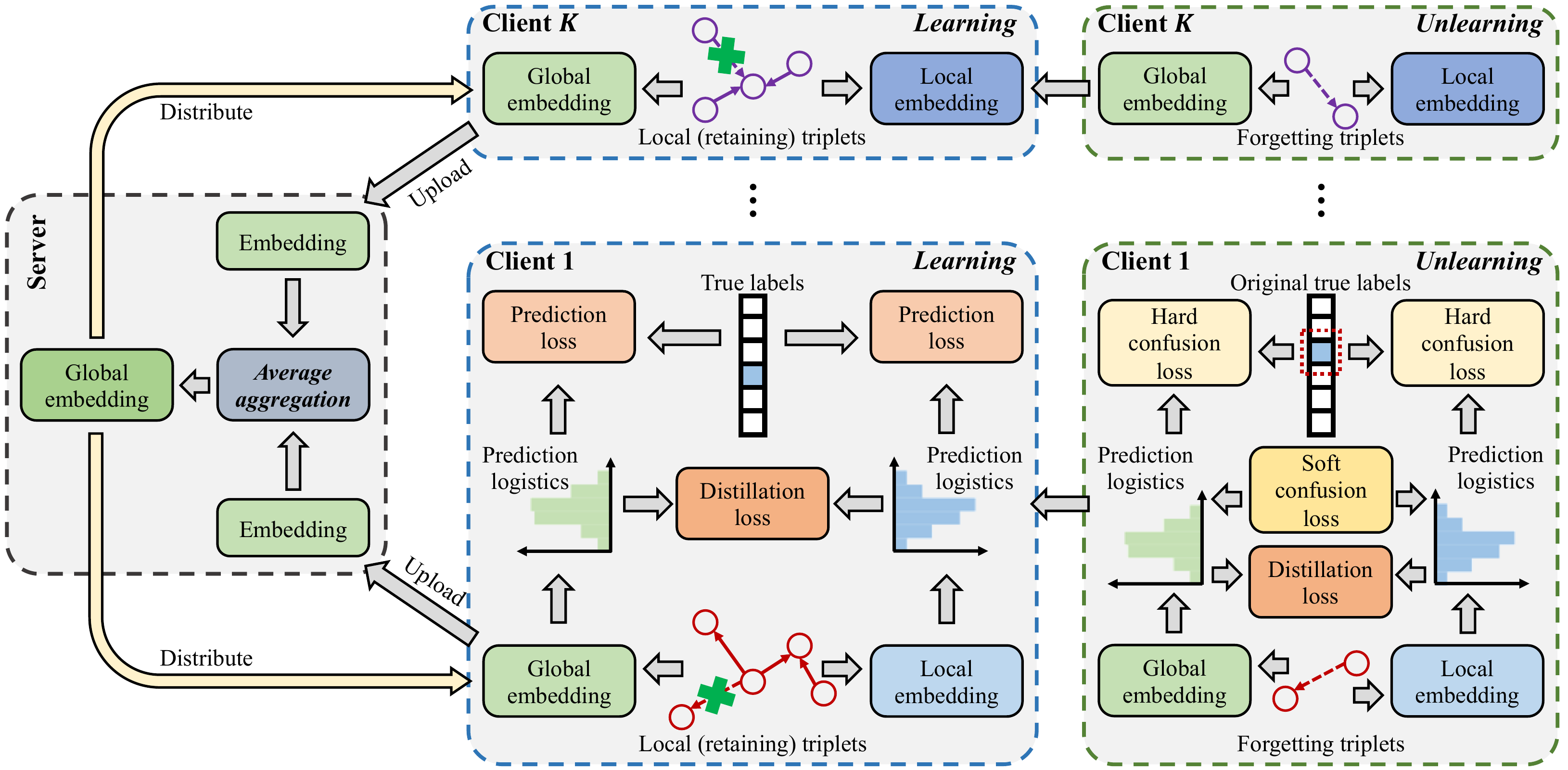}
  \caption{\modelname follows the client-server architecture, and each client contains a learning module and an unlearning module.}
  \label{fig:framework}
\end{figure*}

%--------------------
%\subsection{Overview of \modelname}
%\label{subsect:overview}

We aim to train a KG embedding model based on the FL architecture composed of a central server and a set of clients $K$.
Each client $k\in K$ has a local KG $G^k=\{(h,r,t)\,|\,h,t\in E^k, r\in R^k\}$,  where $E^k,R^k$ denote the entity and relation sets, respectively. 
Local KGs have overlapping entities (i.e., $\forall i, \exists j\neq i, E^i\cap E^j\neq\emptyset$), and the whole dataset is denoted by $G=\bigcup_{k\in K}G^k$.
Furthermore, we do not assume the overlap of relations.
% Local KGs have overlapping entities (i.e., $E^i\cap E^j\neq\emptyset, i\neq j$) but disjoint relations (i.e., $R^i\cap R^j=\emptyset, i\neq j$), and the whole dataset is denoted by $G=\bigcup_{k\in K}G^k$.

At each communication round $t$, the server first samples $K_t\subseteq K$ to collaborate, and then distributes the corresponding local avatar $\mathbf{P}^{k}\mathbf{E}_t$ of global embedding $\mathbf{E}_t$ to each sampled client $k\in K_t$.
Next, the selected client $k$ updates its local embedding $\mathbf{E}^k_{t}$ assisted by $\mathbf{P}^{k}\mathbf{E}_t$ which is received from the server.
At the end of communication round $t$, the server receives the uploaded local avatars of the global embedding and aggregates them into $\mathbf{E}_{t+1}$.
The objective of federated KG embedding learning is to generate a global embedding which minimizes the average local loss:
\begin{align}
\label{eq:fedkgeglobalobjective}
\mathbf{E}=\argmin_{\mathbf{E}}\sum_{k\in K}\frac{|G^k|}{|G|}\mathcal{L}(\mathbf{P}^{k}\mathbf{E};G^k),
\end{align}
where $\mathcal{L}(\cdot)$ is the self-adversarial negative sampling loss of the embedding $\mathbf{P}^{k}\mathbf{E}$ on the local KG $G^k$.

After rounds of communication and training, the framework of \modelname gets the best global embedding $\mathbf{E}$ and local embeddings $\mathbf{E}^k,k\in K$.
Later, the local KG $G^k$ of client $k$ may have a forgetting triplet set $G^k_{u}$ and its complementary triplet set $G^k_{c}$.
To define the goal of unlearning, the local objective is to optimize
\begin{align}
\label{eq:unlearnlocalobjective}
\mathbf{E}^{k-} = \argmin_{\mathbf{E}^k}\Big(-\frac{|G^k_{u}|}{|G^k|}\mathcal{L}(\mathbf{E}^k;G^k_{u})+\frac{|G^k_{c}|}{|G^k|}\mathcal{L}(\mathbf{E}^k;G^k_{c})\Big).
\end{align}

Under the FL setting, a subset of clients $K_{u}\subseteq K$ have their own forgetting sets $\big\{G^{k}_{u}\big\}_{k\in K_{u}}$ and the complementary sets $\big\{G^{k}_{c}\big\}_{k\in K_{u}}$. The global objective of federated unlearning is to obtain a global embedding minimizing the average local loss with unlearning:
\begin{align}
\label{eq:unlearnglobal}
\resizebox{.92\columnwidth}{!}{$
\mathbf{E}^{-}=\argmin\limits_{\mathbf{E}}\sum\limits_{k\in K_{u}}\frac{|G^k|}{|G|}\Big(-\frac{|G^k_{u}|}{|G^k|}\mathcal{L}(\mathbf{P}^{k}\mathbf{E};G^k_{u})+\frac{|G^k_{c}|}{|G^k|}\mathcal{L}(\mathbf{P}^{k}\mathbf{E};G^k_{c})\Big).
$}
\end{align}

As  shown in Figure~\ref{fig:framework}, \modelname serves as a general federated KG embedding learning and unlearning framework for various KG embedding models.
In the learning module of \modelname, we transfer knowledge by mutual knowledge distillation instead of model replacement.
In the unlearning module, we design a two-step method combining retroactive interference and passive decay, which ensures exact forgetting and performance maintenance.
% The unlearning results propagate to the global model through reusing distillation.
% In the unlearning module, we design a two-step method combining retroactive interference and passive decay, which ensures exact forgetting and performance maintenance and propagates well through distillation.

%--------------------
\subsection{Learning in \modelname}
\label{subsect:learn}

For FL algorithms such as FedAvg, clients accept the global model as the initial states of local models for each round of local training, and upload the updated model directly for global aggregation.
The global model is expected to aggregate the knowledge on each client and obtain a balanced performance.
However, FedNTD~\cite{FedNTD} observes that the global convergence and local optimization in FL may interfere with each other, which we call \emph{drift}.
% The global model aggregated from local models may easily lose optimization details of local training.
The global model may easily lose optimization details of local training after aggregation.
Furthermore, the local model tends to forget external knowledge contained in the initial states during training.

Existing work~\cite{MixtureGlobalLocal,ThinkLocallyActGlobally} suggests that there should be a trade-off and separation between global and local models, endowing the FL framework the compatibility of local optimization and global generalization.
Inspired by this, in \modelname we maintain local and global embeddings in parallel that mutually reinforce each other but are not identical.
Besides, identical global and local embeddings may be used to infer knowledge in a private client~\cite{FedR}, resulting in data leakage.
This problem is avoided by separation and communication through mutual distillation in \modelname.

Algorithm~\ref{alg:fedkgedistill} shows the federated KG embedding learning via knowledge distillation of \modelname.
In Line 1, the server first initializes the global embedding model, and then the communication rounds of federated KG embedding learning start.
During each round $t$ (Lines 2--9), a subset of clients $K_t\subseteq K$ are sampled.
The server calculates and distributes corresponding part of the global embedding to each client $k\in K_t$ in Line 5.
Each client refines its local embedding with the global embedding and improves the global embedding with its local embedding through mutual knowledge distillation in parallel in Lines 6--7.
In Line 9, the server aggregates global embeddings without forcing changes to local embeddings, which maintains the coherence of local training and avoids the drift of local optimization.
% $\mathbf{P}^k$ is the permutation matrix from global embedding to local embedding in client $k$ and ${\mathbf{P}^k}^\top$ is its transpose.
% $\mathbf{v}^k$ is an existence vector indicating which entities exist in client $k$.
% $\mathbf{P}^k$ and $\mathbf{v}^k$ can be calculated by the server without sharing raw data through techniques such as private data matching~\cite{PrivateSchemaMatching}.

\begin{algorithm}[!t]
\caption{Federated learning via knowledge distillation}
\label{alg:fedkgedistill}
\KwIn{communication rounds $T$, server, client set $K$, mapping matrix $\mathbf{P}$, existence vector $\mathbf{v}$}
\KwOut{global embedding $\mathbf{E}_T$, local embeddings $\{\mathbf{E}_T^k\}_{k\in K}$}
% \KwOut{global embedding $\mathbf{E}_T$, local embeddings $\{\mathbf{E}_T^k\}_{k=1}^{|K|}$}
Server initializes $\mathbf{E}_0$\;
\For{$t = 0,\dots,T-1$}{
    Sample a client subset $K_t\subseteq K$\;
    \ForAll(\tcp*[f]{run in parallel}){$k\in K_t$}{
        Server sends $\mathbf{E}^k_t=\mathbf{P}^k\mathbf{E}_t$ to client $k$\;
        $\mathbf{E}^{k,\text{local}}_t=\argmin_{\mathbf{E}^k}\mathcal{L}^{\text{local}}(\mathbf{E}^k;\mathbf{E}^k_t,G^k)$\;
        $\mathbf{E}^{k,\text{global}}_t=\argmin_{\mathbf{E}^k}\mathcal{L}^{\text{global}}(\mathbf{E}^k;\mathbf{E}^{k,\text{local}}_t,G^k)$\;
        Client $k$ sends $\mathbf{E}^{k,\text{global}}_t$ to server\;
    }
    $\mathbf{E}_{t+1}=\big(\mathbb{1}\oslash \sum_{k=1}^{K_t}\mathbf{v}^k\big) \otimes\sum_{k\in K_t}{\mathbf{P}^k}^\top\mathbf{E}^{k,\text{global}}_t$\;
}
\end{algorithm}

Let us first dive into how to refine local embeddings with the global embedding by knowledge distillation in \modelname.
Given a triplet $(h,r,t)$, we compute its local score and global score by Eqs. (\ref{eq:localscore}) and (\ref{eq:globalscore}), respectively:
\begin{align}
\label{eq:localscore}
\mathcal{S}^\text{local}_{(h,r,t)}&=\mathcal{S}(h,r,t;\mathbf{E}^\text{local},\mathbf{R}^\text{local}),\\
\label{eq:globalscore}
\mathcal{S}^\text{global}_{(h,r,t)}&=\mathcal{S}(h,r,t;\mathbf{E}^\text{global},\mathbf{R}^\text{local}),
\end{align}
% where $\mathcal{S}(\cdot)$ is the scoring function for particular referred in Section~\ref{subsect:kge}.
where $\mathcal{S}(\cdot)$ is the scoring function referred in Section~\ref{sect:prelim}.
$\mathbf{E}^\text{local}$ and $\mathbf{E}^\text{global}$ are entity embeddings in the local and global models, respectively, and $\mathbf{R}^\text{local}$ is the relation embedding in the local model.
To train each triplet $(h_i,r_i,t_i)\in G^k$, we generate its negative sample set $N(h_i,r_i,t_i)=\{(h_i,r_i,t_{i,j})\,|\,j=1,\dots,n\}$ with size $n$, s.t. $N(h_i,r_i,t_i)\cap G^k=\emptyset$.
We calculate the contrastive prediction loss of $(h_i,r_i,t_i)$ as follows:
\begin{align}
\label{eq:localpredictloss}
\resizebox{.92\columnwidth}{!}{$
\mathcal{L}^\text{predict}_{(h_i,r_i,t_i)}= -\log\Big(\sigma\big(\mathcal{S}^\text{local}_{(h_i,r_i,t_i)}\big)\Big) -\quad\smashoperator{\sum_{(h,r,t)\in N(h_i,r_i,t_i)}}\ \frac{1}{n}\log\Big(\sigma\big(-\mathcal{S}^\text{local}_{(h,r,t)}\big)\Big),
$}
\end{align}
where $\sigma(\cdot)$ is the sigmoid activation function.

To avoid the inconsistency in the optimization direction of local and global embeddings, knowledge distillation is conducted along with sample prediction.
The score of a KG embedding model on a triplet characterizes the probability of being predicted as positive.
So, we transfer knowledge by distilling the distribution of local and global scores on samples and their negative sets.
For a triplet $(h_i,r_i,t_i)$, we compute its distillation loss for the local embedding:
\begin{align}
\label{eq:localdistillloss}
\mathcal{L}^\text{distill}_{(h_i,r_i,t_i)}=\text{KL}\Big(\mathcal{P}^\text{local}_{(h_i,r_i,\cdot)},\mathcal{P}^\text{global}_{(h_i,r_i,\cdot)}\Big),
\end{align}
where $\text{KL}(\cdot)$ is the Kullback-Leiber distillation function.
% $\mathcal{P}^\text{local}_{(h_i,r_i,\cdot)}$ is the score distribution of sample $(h_i,r_i,t_i)$ generated by uniting softmax of positive sample score $\mathcal{P}^\text{local}_{(h_i,r_i,t_i)}$ and softmax of negative sample score $\mathcal{P}^\text{local}_{(h_i,r_i,t_{i,j})}$, which are computed as follows:
$\mathcal{P}^\text{local}_{(h_i,r_i,\cdot)}$ is the local score distribution of sample $(h_i,r_i,t_i)$ generated by combining $\mathcal{P}^\text{local}_{(h_i,r_i,t_i)}$ and $\mathcal{P}^\text{local}_{(h_i,r_i,t_{i,j})}$, which are computed as follows:
\begin{align}
\label{eq:localposprob}
\mathcal{P}^\text{local}_{(h_i,r_i,t_i)}=\frac{\exp\big(s^\text{local}_{(h_i,r_i,t_i)}\big)}{\sum_{(h,r,t)\in N(h_i,r_i,t_i)\cup\{(h_i,r_i,t_i)\}}\exp\big(s^\text{local}_{(h,r,t)}\big)},\\
\label{eq:localnegprob}
\mathcal{P}^\text{local}_{(h_i,r_i,t_{i,j})}=\frac{\exp\big(s^\text{local}_{(h_i,r_i,t_{i,j})}\big)}{\sum_{(h,r,t)\in N(h_i,r_i,t_i)\cup\{(h_i,r_i,t_i)\}}\exp\big(s^\text{local}_{(h,r,t)}\big)},
\end{align}
and $\mathcal{P}^\text{global}_{(h_i,r_i,\cdot)}$ is the global score distribution of $(h_i,r_i,t_i)$ generated by combining $\mathcal{P}^\text{global}_{(h_i,r_i,t_i)}$ and $\mathcal{P}^\text{global}_{(h_i,r_i,t_{i,j})}$, which can be calculated in a similar way.

Finally, the local model of client $k$ is equipped with the two jointly optimized losses as follows:
\begin{align}
\label{eq:localloss}
\mathcal{L}^\text{local} = \sum_{(h_i,r_i,t_i)\in G^k}\Big(\mathcal{L}^\text{predict}_{(h_i,r_i,t_i)}+\mu_\text{distill}\,\mathcal{L}^\text{distill}_{(h_i,r_i,t_i)}\Big),
\end{align}
where $\mu_\text{distill}$ is the parameter to adjust the degree of distillation.

The joint loss $\mathcal{L}^\text{global}$ to optimize global embedding $\mathbf{E}^\text{global}$ can be computed similarly. 

\smallskip
\noindent\textbf{Complexity analysis.}
Given total communication round $T$, client number $K$, average entity number $N'$, total entity number $N$, total relation number $M$, triple number $P$ and hidden dimension $H$, \modelname's parameter complexity is $(KN'+N+M)H$, computation complexity is $\mathcal{O}(PHT)$, and total communication cost is $2KN'HT$.

%--------------------
\subsection{Unlearning in \modelname}
\label{subsect:unlearn}

To solve the difficulty in federated KG embedding unlearning, we resort to the forgetting theories in cognitive neuroscience.
There are two major theories explaining for forgetting of cognition, namely interference and decay.
The interference theory~\cite{retroactiveinterference} posits that forgetting occurs when memories compete and interfere with others.
The decay theory~\cite{decay} believes that memory traces fade and disappear, and eventually lost if not retrieved and rehearsed.
We propose a two-step unlearning method for federated KG embedding.
Inspired by the interference theory, \modelname first conducts a retroactive interference step with hard and soft confusions.
Then, according to the decay theory, \modelname performs a passive decay step, which can recover the performance loss caused by interference while suppressing the activation of the  forgotten knowledge.

For client $k$ to perform unlearning with its local KG $G^k$, $G^k$ is split into the forgetting set $G^k_u$ and the retaining set $G^k_c$.
We first introduce the retroactive interference step conducted to the local embedding.
To unlearn each triplet $(h_i,r_i,t_i)\in G^k_u$, we have its negative sample set $N(h_i,r_i,t_i)=\{(h_i,r_i,t_{i,j})\,|\,j=1,\dots,n\}$ with size $n$, s.t. $N(h_i,r_i,t_i)\cap G^k=\emptyset$.
We first compute the hard confusion loss to optimize the local embedding by treating $(h_i,r_i,t_i)$ as negative:
\begin{align}
\label{eq:unlearnhard}
\resizebox{.9\columnwidth}{!}{$
\mathcal{L}^\text{hard}_{(h_i,r_i,t_i)}=-\log\Big(\sigma\big(-\mathcal{S}^\text{local}_{(h_i,r_i,t_i)}\big)\Big) -\quad\smashoperator{\sum_{(h,r,t)\in N(h_i,r_i,t_i)}}\ \frac{1}{n}\log\Big(\sigma\big(-\mathcal{S}^\text{local}_{(h,r,t)}\big)\Big).
$}
\end{align}

% Empirically~\cite{federated_unlearning_class_discrim}, the unlearned models should be compensated for the accuracy degradation after unlearning.
Unfortunately, if we blindly optimize the forgetting sets as negative, the embeddings are likely to be polluted.
In fact, unlearning may leave trace of the forgetting set and thus results in privacy risks such as membership inference attacks~\cite{unlearn_jeopardize_privacy}.
So, we design soft confusion, which reflects the distance between scores of the triplet in the forgetting set and its negative samples.
By minimizing the soft confusion, the scores of the triplet and negatives are forced closer, which can help not only forgetting the triplet, but also preventing the overfitting of unlearning.
The soft confusion loss for the triplet $(h_i,r_i,t_i)$ is calculated as follows:
\begin{align}
\label{eq:unlearnsoft}
\mathcal{L}^\text{soft}_{(h_i,r_i,t_i)}=\sum_{(h,r,t)\in N(h_i,r_i,t_i)}\frac{1}{n}\,\Big|\Big|\,\mathcal{S}^\text{local}_{(h,r,t)}-\mathcal{S}^\text{local}_{(h_i,r_i,t_i)}\,\Big|\Big|_{L2}.
\end{align}

To perform the retroactive interference while keeping the relevance between local and global embeddings, we calculate the total interference loss combining the KL divergence loss in Eq.~(\ref{eq:localdistillloss}) as
\begin{align}
\label{eq:localunlearn}
\resizebox{.9\columnwidth}{!}{$
\mathcal{L}^\text{interference}_{(h_i,r_i,t_i)}=\mathcal{L}^\text{hard}_{(h_i,r_i,t_i)}+\mu_\text{soft}\,\mathcal{L}^\text{soft}_{(h_i,r_i,t_i)}+\mu_\text{distill}\,\mathcal{L}^\text{distill}_{(h_i,r_i,t_i)}.
$}
\end{align}

The interference loss of the global embedding is calculated similarly, by changing the local scores to the global scores and reversing the distillation.

After the retroactive interference step, memories of the global and local embeddings on the forgetting set are erased.
However, a significant decrease in model performance can happen.
On one hand, the model optimization in the retroactive interference is limited to specific triplets to forget, which slightly destroys the generalization of the embedding model.
On the other hand, the unlearning of a specific triplet would affect its associated triplets.
So, we perform a passive decay step afterwards by mutual knowledge distillation with $G^k_c$ as input.
$\mathbf{E}^{k,global}$ and $\mathbf{E}^{k,local}$ are optimized alternately in batches as the teacher model to each other.
Learning on the retaining set can recover the generalization of the model.
Mutual distillation suppresses the activation of the forgetting triplets.

%==============================
\section{Experiments and Results}
\label{sect:exp}

We implemented our framework, \modelname, on a server with four Intel Xeon Gold 6326 CPUs, 512GB memory and four NVIDIA RTX A6000 GPUs. 
In this section, we first introduce our newly-constructed datasets for federated KG embedding.
Then, we carry out experiments to evaluate \modelname on federated KG embedding learning and unlearning.
The datasets and source code are online.\footnote{\url{https://doi.org/10.5281/zenodo.7601676}}

%--------------------
\subsection{Clustering-based Dataset Construction}
\label{subsect:dataset}

To evaluate the effectiveness of \modelname, we select the benchmark KG embedding dataset FB15k-237 as the original dataset.
Following the experiment setting of FedE~\cite{FedE}, we averagely partition relations and distribute triplets into three, five or ten clients accordingly, forming three datasets called FB15k-237-R3, R5 and R10.
These datasets satisfy the non-IID condition with no overlapping relations.
However, they do not conform to the entity distribution and graph structure in real-world federation, because relations and entities on each client may be unrelated.

In this paper, we propose a new dataset construction algorithm, which first partitions relations by clustering and then distributes triplets into clients.
This generates more realistic datasets for federated KG embedding.
Due to the space limitation, we present the algorithm in Appendix~\ref{app:algo}.
In line with random partitioning, we name the new heterogeneous datasets FB15k-237-C3, C5 and C10.

\begin{figure}
  \centering
  \includegraphics[width=\linewidth]{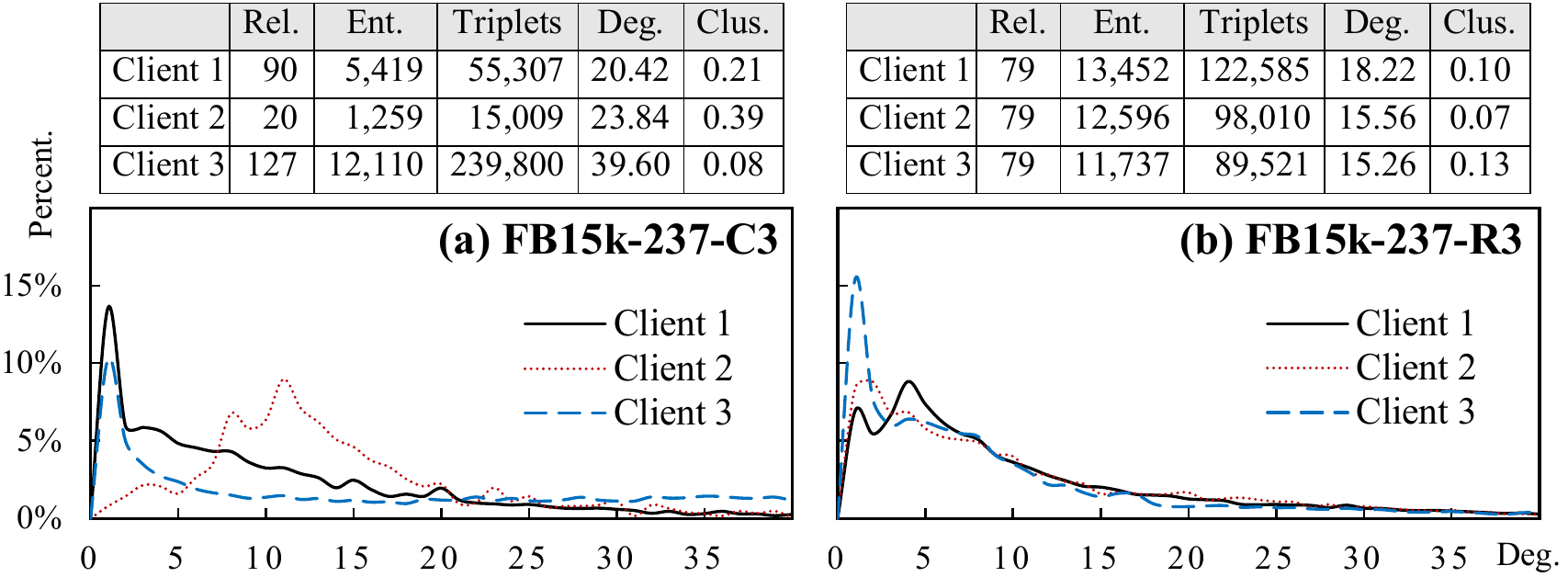}
  \caption{Comparison of degree distributions and clustering coefficients between FB15k-237-C3 and FB15k-237-R3.}
  \label{fig:degree1}
\end{figure}

We compare the two types of datasets constructed with random partition and clustering-based partition.
Figure~\ref{fig:degree1} shows the comparison results of FB15k-237-C3 and FB15k-237-R3, and those between FB15k-237-C5/C10 and FB15k-237-R5/R10 are supplemented in Appendix~\ref{app:dataset}.
We can see that FB15k-237-C3 exhibits more differences in the relation, entity and triplet numbers.
The average degrees and clustering coefficients of local datasets in FB15k-237-C3 are larger and more varying than those in FB15k-237-R3.
Furthermore, the numbers of overlapping entities of FB15k-237-C3 and R3 are 1,447 and 11,067, respectively.
This shows two key properties of our clustering-based datasets: (i) the local data distribution of each client is smooth, and the agglomeration is strong; and (ii) the data among clients are highly heterogeneous.

%--------------------
\subsection{Experiments on Federated Learning}

\begin{table*}[!t]
\centering
\caption{Average link prediction results on FB15k-237-C3/C5/C10}
{\small
\begin{tabular}{ll|cccc|cccc|cccc}
\toprule 
	\multicolumn{2}{c|}{\multirow{2}{*}{Methods}} & \multicolumn{4}{c|}{FB15k-237-C3} & \multicolumn{4}{c|}{FB15k-237-C5} & \multicolumn{4}{c}{FB15k-237-C10} \\
	\cmidrule(lr){3-6} \cmidrule(lr){7-10} \cmidrule(lr){11-14}
	& & Hits@1 & Hits@3 & Hits@10 & MRR & Hits@1 & Hits@3 & Hits@10 & MRR & Hits@1 & Hits@3 & Hits@10 & MRR \\
\midrule
	\parbox[t]{3mm}{\multirow{7}{*}{\rotatebox[origin=c]{90}{TransE}}} & Independent & 19.48 & 37.57 & 53.91 & 31.44 & 19.77 & 39.77 & 56.28 & 32.60 & 18.68 & 38.56 & 55.66 & 31.52 \\
	& Centralized & 20.40 & 38.79 & 56.14 & 32.65 & 20.53 & 38.98 & 56.53 & 32.86 & 20.17 & 38.57 & 56.29 & 32.49 \\
	& FedE & 20.25 & 39.19 & 56.18 & 32.66 & 20.66 & 39.78 & 57.12 & 33.23 & 19.86 & 38.15 & 55.82 & 32.10 \\
	& FedProx & 20.20 & 38.76 & 55.91 & 32.50 & 21.02 & \underline{39.99} & 57.36 & 33.57 & 20.04 & 38.62 & 56.17 & 32.43 \\
	& FedEC & 20.65 & 39.36 & 56.44 & 32.99 & 21.11 & 39.90 & 57.42 & 33.54 & 20.05 & 38.67 & 56.27 & 32.42 \\
	& \modelname (local) & \textbf{21.27} & \textbf{40.21} & \textbf{57.53} & \textbf{33.71} & \textbf{21.89} & \textbf{40.58} & \textbf{58.19} & \textbf{34.32} & \textbf{21.05} & \textbf{39.66} & \textbf{57.44} & \textbf{33.44} \\
	& \modelname (global) & \underline{20.94} & \underline{39.71} & \underline{56.89} & \underline{33.30} & \underline{21.38} & \textbf{40.58} & \underline{57.96} & \underline{33.98} & \underline{20.53} & \underline{39.21} & \underline{56.82} & \underline{32.95} \\
\midrule
	\parbox[t]{3mm}{\multirow{7}{*}{\rotatebox[origin=c]{90}{ComplEx}}} & Independent & 23.85 & 39.10 & 54.40 & 34.21 & 22.90 & 37.47 & 52.21 & 32.87 & 23.52 & 39.20 & 54.46 & 34.05 \\
	& Centralized & \underline{23.95} & 40.51 & 56.45 & 35.02 & \textbf{24.47} & \underline{40.96} & \underline{57.22} & \underline{35.57} & \underline{24.73} & \underline{40.70} & \underline{56.69} & \underline{35.52} \\
	& FedE & 20.88 & 36.22 & 52.31 & 31.45 & 19.82 & 35.32 & 51.65 & 30.49 & 19.52 & 34.79 & 50.61 & 30.02 \\
	& FedProx & 22.67 & 41.04 & 57.82 & 34.71 & 22.75 & 38.54 & 54.88 & 33.57 & 20.73 & 38.72 & 55.70 & 32.66 \\
	& FedEC & 22.44 & 40.97 & 57.91 & 34.58 & 19.63 & 37.06 & 54.09 & 31.30 & 20.59 & 38.36 & 55.65 & 32.53 \\
	& \modelname (local) & \textbf{25.38} & \textbf{43.37} & \textbf{59.46} & \textbf{37.10} & \underline{24.05} & \textbf{41.74} & \textbf{58.11} & \textbf{35.66} & \textbf{25.14} & \textbf{42.01} & \textbf{57.92} & \textbf{36.33} \\
	& \modelname (global) & 23.58 & \underline{41.50} & \underline{57.94} & \underline{35.37} & 22.33 & 39.95 & 56.81 & 34.07 & 22.12 & 38.92 & 55.40 & 33.45 \\
\midrule
	\parbox[t]{3mm}{\multirow{7}{*}{\rotatebox[origin=c]{90}{RotatE}}} & Independent & 24.31 & 43.13 & 59.55 & 36.47 & 26.43 & 43.36 & 59.45 & 37.64 & 23.94 & 41.80 & 57.95 & 35.67 \\
	& Centralized & 25.68 & 44.55 & 61.55 & 37.94 & 26.23 & 45.05 & 62.05 & 38.48 & 25.29 & 44.39 & \textbf{61.64} & 37.76 \\
	& FedE & 21.23 & 39.81 & 56.71 & 33.46 & 21.94 & 40.70 & 57.62 & 34.24 & 20.54 & 38.96 & 56.41 & 32.82 \\
	& FedProx & 26.07 & 45.01 & 61.71 & 38.33 & 26.92 & \textbf{45.75} & \textbf{62.49} & \underline{39.14} & \underline{25.75} & \underline{44.52} & 61.53 & 37.99 \\
	& FedEC & 26.13 & 45.20 & 61.76 & 38.38 & 26.50 & 45.12 & 61.82 & 38.67 & 25.53 & 44.49 & 61.33 & 37.86 \\
	& \modelname (local) & \underline{26.80} & \underline{45.50} & \textbf{62.13} & \textbf{38.92} & \textbf{27.52} & 45.56 & 62.21 & \textbf{39.34} & \textbf{26.01} & \textbf{44.83} & \underline{61.63} & \textbf{38.22} \\
	& \modelname (global) & \textbf{26.81} & \textbf{45.56} & \underline{61.93} & \underline{38.90} & \underline{27.08} & \underline{45.57} & \underline{62.24} & 39.13 & 25.71 & \textbf{44.83} & 61.32 & \underline{38.00} \\
\bottomrule
    \multicolumn{14}{l}{The best and second best scores are marked in \textbf{bold} and with \underline{underline}, respectively.}
\end{tabular}}
\label{tab:lp_fb237c}
\end{table*}

\begin{table*}[!t]
\centering
\caption{Average unlearning results on different sets of FB15K-237-C3 with the local and global models}
{\small
\begin{tabular}{l|cccccc|cccccc}
\toprule 
	\multirow{2}{*}{Hits@1} & \multicolumn{2}{c}{Raw local} & \multicolumn{2}{c}{Re-trained local} & \multicolumn{2}{c|}{Unlearned local} & \multicolumn{2}{c}{Raw global} & \multicolumn{2}{c}{Re-trained global} & \multicolumn{2}{c}{Unlearned global}\\
	\cmidrule(lr){2-3} \cmidrule(lr){4-5} \cmidrule(lr){6-7} \cmidrule(lr){8-9} \cmidrule(lr){10-11} \cmidrule(lr){12-13} 
	& Forget$\downarrow$ & Test$\uparrow$ & Forget$\downarrow$ & Test$\uparrow$ & Forget$\downarrow$ & Test$\uparrow$ & Forget$\downarrow$ & Test$\uparrow$ & Forget$\downarrow$ & Test$\uparrow$ & Forget$\downarrow$ & Test$\uparrow$ \\
\midrule
	TransE & 38.86 & 21.27 & 21.10 & 20.86 & 13.23 & 21.04 & 36.83 & 20.88 & 20.22 & 19.94 & 15.02 & 20.95 \\
	ComplEx & 59.16 & 25.38 & 17.60 & 16.70 & \ \ 6.26 & 23.49 & 54.05 & 24.13 & 25.63 & 23.98 & \ \ 5.72 & 23.03 \\
	RotatE & 64.54 & 26.80 & 25.17 & 25.87 & 19.65 & 26.66 & 62.99 & 26.86 & 26.46 & 25.78 & 22.73 & 26.24 \\
\bottomrule
\toprule 
	\multirow{2}{*}{MRR} & \multicolumn{2}{c}{Raw local} & \multicolumn{2}{c}{Re-trained local} & \multicolumn{2}{c|}{Unlearned local} & \multicolumn{2}{c}{Raw global} & \multicolumn{2}{c}{Re-trained global} & \multicolumn{2}{c}{Unlearned global}\\
	\cmidrule(lr){2-3} \cmidrule(lr){4-5} \cmidrule(lr){6-7} \cmidrule(lr){8-9} \cmidrule(lr){10-11} \cmidrule(lr){12-13} 
	& Forget$\downarrow$ & Test$\uparrow$ & Forget$\downarrow$ & Test$\uparrow$ & Forget$\downarrow$ & Test$\uparrow$ & Forget$\downarrow$ & Test$\uparrow$ & Forget$\downarrow$ & Test$\uparrow$ & Forget$\downarrow$ & Test$\uparrow$ \\
\midrule
	TransE & 52.54 & 33.71 & 33.23 & 33.03 & 24.56 & 33.43 & 50.71 & 33.32 & 32.33 & 32.18 & 27.09 & 33.33\\
	ComplEx & 70.44 & 37.10 & 27.32 & 26.53 & 13.83 & 35.14 & 65.57 & 35.67 & 37.48 & 36.13 & 12.70 & 34.76 \\
	RotatE & 74.87 & 38.92 & 37.57 & 37.82 & 32.47 & 38.73 & 73.71 & 38.98 & 38.42 & 37.79 & 36.40 & 38.36 \\
\bottomrule
\end{tabular}
}
\label{tab:unlearn_fb237c3}
\end{table*}

\noindent\textbf{Baselines.} 
To validate the effectiveness of learning in \modelname, we compare \modelname with various federated KG embedding methods, including FedE~\cite{FedE}, FedProx~\cite{FedProx}\footnote{Our implementation of FedProx is FedE with a regularization term, which keeps local embeddings close to global embeddings by minimizing their distance $||\mathbf{E}^k-\mathbf{E}^k_t||_{L2}$.}, FedEC~\cite{FedEC}, and two baseline settings, namely independent and centralized.
The independent setting means training separate embeddings for each client with its local dataset, and the centralized setting means training a global embedding by aggregating the local datasets from all clients.
All these methods and settings are realized with three representative KG embedding models, i.e., TransE~\cite{TransE}, ComplEx~\cite{ComplEx} and RotatE~\cite{RotatE}.

\smallskip
\noindent\textbf{Evaluation setup.} 
Following the convention, we use average Hits@$N$ ($N=1,3,10$) and mean reciprocal rank (MRR) of all clients under the filtered setting to evaluate the entity link prediction performance of each method.
The local triplets in each client are randomly divided with the ratio 8:1:1 for training, validation and test.
We report the results of local and global embeddings of \modelname.
% We report the average prediction results of each client with local and global embeddings of \modelname.
% We report both the prediction results of the test sets on the local and global embeddings of \modelname.

\smallskip
\noindent\textbf{Implementation details.} 
% The results of global embeddings are obtained by testing after the aggregated global embedding is mapped to each client.
We employ the Adam~\cite{Adam} optimizer with a learning rate of 1e-4.
% For the independent and centralized settings, we evaluate on validation per 5 epochs and early stop patience is employed with 3 epochs.
For the independent and centralized settings, we evaluate the validation set every 5 epochs. 
We further use early stop based on MRR on the validation sets with a patience of 3.
% For \modelname and all competitors, we set local training epoch to 3 per communication round, evaluate on validation per 5 rounds and early stop patience is employed with 3 rounds.
For \modelname and other competitors, we train the local model for 3 epochs in each communication round.
We also evaluate the validation sets every 5 communication rounds.
Moreover, we set the patience to 3 for early stop based on the average MRR of all clients.
% Given the limited number of clients, the sampling fraction to select involved clients in each round of federated KG embedding learning is set to 1.0.
% Given that the number of clients is limited to 3/5/10 in our problem setting, the sampling fraction to select involved clients in each round of federated KG embedding learning is set to 100\% following FedEC.
The sampling fraction to select involved clients in each round of federated KG embedding learning is set to 100\% following FedE and FedEC.
Note that \modelname has the ability to deal with varying sampling fractions.
For the KG embedding models, we set the batch size to 1,024, the negative sampling size to 256, and the dimension of entity and relation embeddings to 256.
For FedProx, we set $\mu$ for the proximal term to 0.1.
For FedEC, we set $\tau$ to 0.2 and $\mu_\text{con}$ to 0.3.
For \modelname, we set $\mu_{\text{distill}}$ to 2.

\smallskip
\noindent\textbf{Results.} 
The results on FB15k-237-C3, C5 and C10 are presented in Table~\ref{tab:lp_fb237c}, and those on FB15k-237-R3, R5 and R10 are shown in Appendix~\ref{app:lp_fb237_R}.
We can find that (i) our proposed \modelname can generally achieve better performance in link prediction than others.
(ii) For different client numbers, \modelname achieves consistent performance after convergence.
As shown in Figure~\ref{fig:conv} shortly, the convergence round becomes larger with the number of clients increases.
(iii) Compared to the independent setting, FedProx, FedEC and \modelname all gain improvement, while the results of FedE do not significantly increase or even decline sometimes.
This indicates that FedE can hardly handle such heterogeneous datasets.
(iv) We also see that, in several cases, \modelname even outperforms the centralized setting, suggesting that for the datasets with strong heterogeneity, smoothly exchanging knowledge through mutual distillation is more appropriate than directly sharing data.
% (iv) In most cases, \modelname (global) achieves comparable performance, which is different from that of \modelname (local).
(v) In most cases, \modelname (global) achieves the second-best performance, which is only behind \modelname (local).
This shows that \modelname obtains a global entity embedding which is compatible with local embeddings in clients.

%--------------------
\subsection{Experiments on Federated Unlearning}

\noindent\textbf{Baselines.} 
To assess the federated unlearning in \modelname, we compare the performance of raw, re-trained and unlearned embeddings.
The raw embeddings are derived from federated learning of \modelname.
The re-trained embeddings are re-trained from scratch using the retaining triplets.
The unlearned embeddings are generated from raw embeddings through unlearning in \modelname.
% To measure the performance of federated unlearning in \modelname, we use the re-trained local and global embeddings for comparison.
% The re-trained local embeddings are obtained by re-training a single embedding model for each client with their local retained triplets.
% In the re-trained global setting, the local retained KGs in memory are collected for re-training a new global model.

\smallskip
\noindent\textbf{Evaluation setup.} 
We randomly sample a forgetting set from the original training set of each client with a proportion of 1\%.
We feed the forgetting sets into re-trained/unlearned embeddings, and compare the prediction results with those obtained by the raw embeddings, to measure the ability of \modelname in forgetting specific knowledge.
Smaller Hits@1 and MRR on the forgetting sets indicate that the model forgets more thoroughly.
% We feed the forgetting sets into re-trained/unlearned embeddings as test sets, and compare the output prediction with the raw embeddings, to measure the ability of \modelname in forgetting specific knowledge.
In addition, we use the original test sets to quantify how much performance reduction that we have to pay as the cost of unlearning.
We compare the local and global embeddings to see if the unlearned knowledge can be propagated to the global embedding smoothly in \modelname.
The re-trained local embeddings are obtained with a single embedding model for each client with their local retaining sets.
The re-trained global embedding is generated by aggregating the retaining sets of all clients.
We report the average Hits@1 and MRR scores on the forgetting and test sets of all clients.
% We also present the results of the raw and re-trained local and global embeddings for comparison.

\smallskip
\noindent\textbf{Implementation details.} 
% For re-training local and global embeddings, we evaluate on validation per 5 epochs and early stop patience is set to 3 epochs.
% The entity and relation embedding dimension of re-trained models is set to 256.
% The re-trained models are optimized by Adam with a learning rate of 1e-4.
% We set the batch size to 1024 and negative sampling size to 256, which is same to \modelname.
For the re-trained embeddings, we follow the previous hyperparameter settings.
For the retroactive interference in unlearning, we set $\mu_{\text{soft}}$ to 0.1.
Moreover, we set the training epoch of unlearning to 10, which is much smaller than re-training.

\smallskip
\noindent\textbf{Results.} 
The unlearning results on the FB15k-237-C3 dataset are shown in Table~\ref{tab:unlearn_fb237c3}, and other results are given in Appendix \ref{app:unlearn_fb237_C5C10}.
We have three findings:
(i) The Hits@1 and MRR scores of the re-trained models on the forgetting sets are close to those on the test sets.
This indicates that the re-trained models are still able to predict missing links, i.e., completely forgetting cannot be achieved through re-training from scratch.
(ii) The scores of the unlearned embeddings drop to a lower level on the forgetting sets than the raw embeddings and the re-trained embeddings, suggesting that \modelname is able to erase the knowledge thoroughly and suppress the activation of the memory to the forgetting sets during unlearning.
(iii) The unlearned embeddings achieve comparable performance on the test sets with the raw models, which is higher than the re-trained models.
This indicates that unlearning in \modelname can maintain the global knowledge absorbed in FL.
% It can be found that, the Hits@1 and MRR scores which are obtained by the unlearned embeddings drop to a quite low level on the forgetting sets.
% Compared to the re-trained local and global embeddings, the unlearned local and global embeddings have lower Hits@1 and MRR on the forgetting sets, and achieve comparable performance on the test sets.
% We can obtain the following findings based on the above observations:
% (i) The re-trained embeddings are still able to predict missing links on the forgetting sets.
% However, \modelname is able to suppress the activation of the memory to the forgetting sets during unlearning;
% (ii) \modelname can maintain the global knowledge absorbed in FL during the unlearning process, thus achieving better performance than the re-trained local embeddings.

%--------------------
\subsection{Further Analysis}

\noindent\textbf{Ablation study.}
Regarding the learning in \modelname, if we abrogate the mutual knowledge distillation between the global and local embeddings, the global embedding degrades to the entity embedding in FedE and the local embeddings degrade to the entity embeddings in the independent setting.
The performance of the global and local embeddings in \modelname is much better than FedE and the independent setting, which validates the smooth knowledge exchange realized by mutual knowledge distillation.

To evaluate the effectiveness of the hard and soft confusions in the retroactive interference step of \modelname, we modify two variants in federated unlearning, namely \modelname without hard confusion and \modelname without soft confusion.
From Table~\ref{tab:unlearn_fb237c3_ablation}, we find that:
(i) The Hits@1 scores of \modelname without hard or soft confusion on the forgetting sets are much lower than those of the raw models in Table~\ref{tab:unlearn_fb237c3}, suggesting that both of them are effective.
(ii) \modelname without hard confusion shows higher Hits@1 on the forgetting sets, indicating that the hard confusion plays a key role in unlearning.
(iii) The Hits@1 scores of \modelname without soft confusion is slightly higher than the original \modelname.
This shows that the soft confusion acts as a fine-tuning factor in retroactive inference.
% As shown in Table~\ref{tab:unlearn_fb237c3_ablation}, \modelname without hard confusion has a higher Hits@1 on the forgetting sets, which implies that the hard confusion plays an important role in unlearning.
% We also find that \modelname without soft confusion has a slightly higher Hits@1 on the forgetting sets and lower Hits@1 on the test sets.
% This verifies that the soft confusion acts as a fine-tune factor in retroactive interference, which makes samples in the forgetting set more difficult to predict.

\begin{table}[!t]
\centering
\caption{Ablation studies on FB15K-237-C3}
{\small
\begin{tabular}{l|cc|cc}
\toprule 
	\multirow{2}{*}{Hits@1} & \multicolumn{2}{c|}{Unlearned local} & \multicolumn{2}{c}{Unlearned global} \\
	\cmidrule(lr){2-3} \cmidrule(lr){4-5}  
	& Forget$\downarrow$ & Test$\uparrow$ & Forget$\downarrow$ & Test$\uparrow$ \\
\midrule
	TransE & 13.23 & 21.04 & 15.02 & 20.95 \\
	\quad w/o hard confusion & 17.32 & 21.18 & 18.06 & 20.99 \\
	\quad w/o soft confusion & 15.87 & 20.73 & 15.28 & 20.59 \\
\midrule
	ComplEx & \ \ 6.26 & 23.49 & \ \ 5.72 & 23.03 \\
	\quad w/o hard confusion & 19.94 & 23.31 & 18.36 & 22.82 \\
	\quad w/o soft confusion & \ \ 7.71 & 23.36 & \ \ 6.48 & 23.00 \\
\midrule
	RotatE & 19.65 & 26.66 & 22.73 & 26.24 \\
	\quad w/o hard confusion & 34.23 & 26.06 & 32.60 & 25.84 \\
	\quad w/o soft confusion & 21.59 & 26.23 & 24.69 & 26.24 \\
\bottomrule
\end{tabular}
}
\label{tab:unlearn_fb237c3_ablation}
\end{table}

\begin{figure}[!t]
  \centering
  \includegraphics[width=\linewidth]{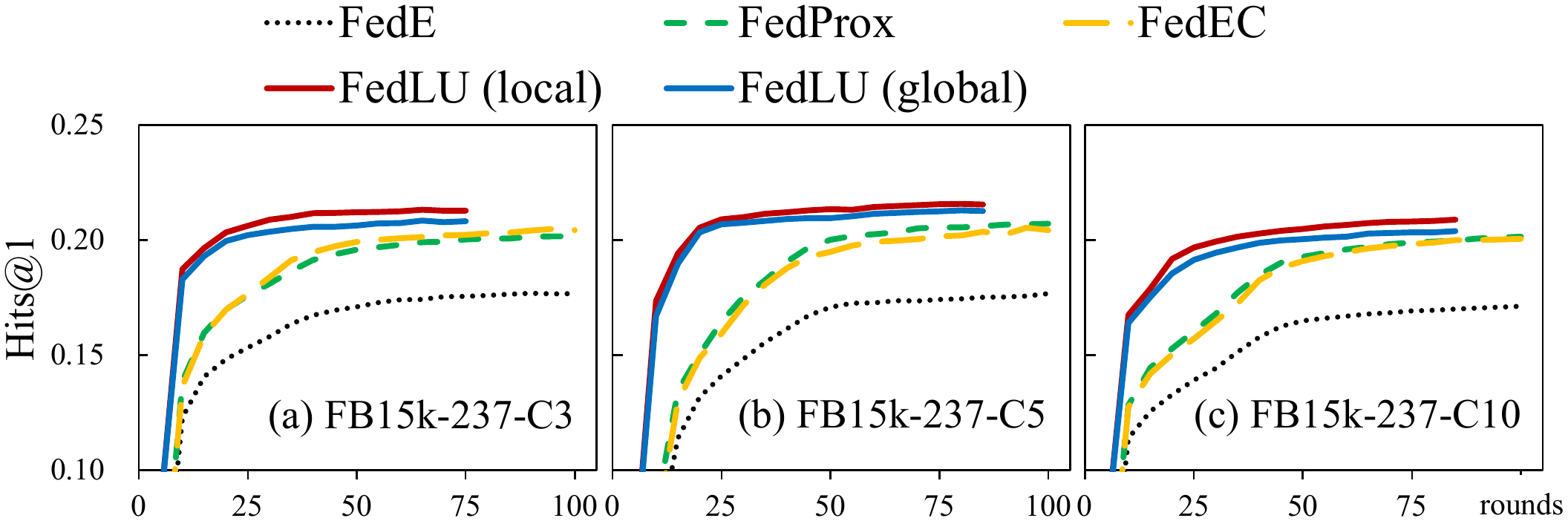}
  \caption{Hits@1 results of TransE on the validation sets w.r.t. communication rounds.}
  \label{fig:conv}
\end{figure}

\smallskip
\noindent\textbf{Convergence speed.}
Figure~\ref{fig:conv} shows the average Hits@1 of \modelname (TransE) on the validation sets w.r.t. communication rounds, compared with other competitors.
In addition to the higher average Hits@1, we can also see that the mutual distillation of the local and global embeddings makes \modelname converge faster, while other competitors converge slower due to directly using the aggregated global embedding for initializing local training.

\begin{figure}[!t]
\centering
\subfigure[FedE]{
\label{fig:tsne_TransE_c3_FedE} %
\includegraphics[height = 30 mm]{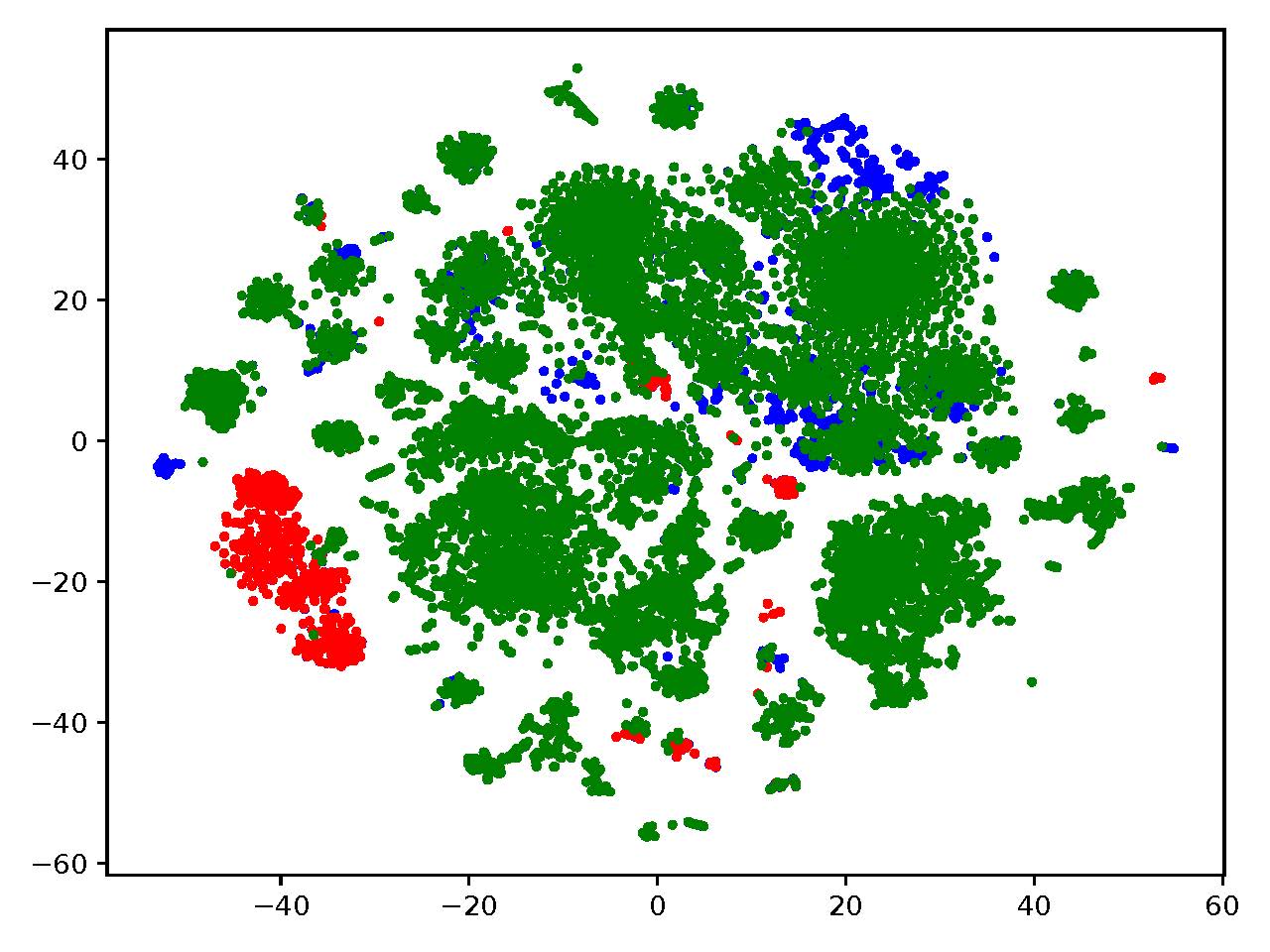}}
\subfigure[FedProx]{
\label{fig:tsne_TransE_c3_FedProx} %
\includegraphics[height = 30 mm]{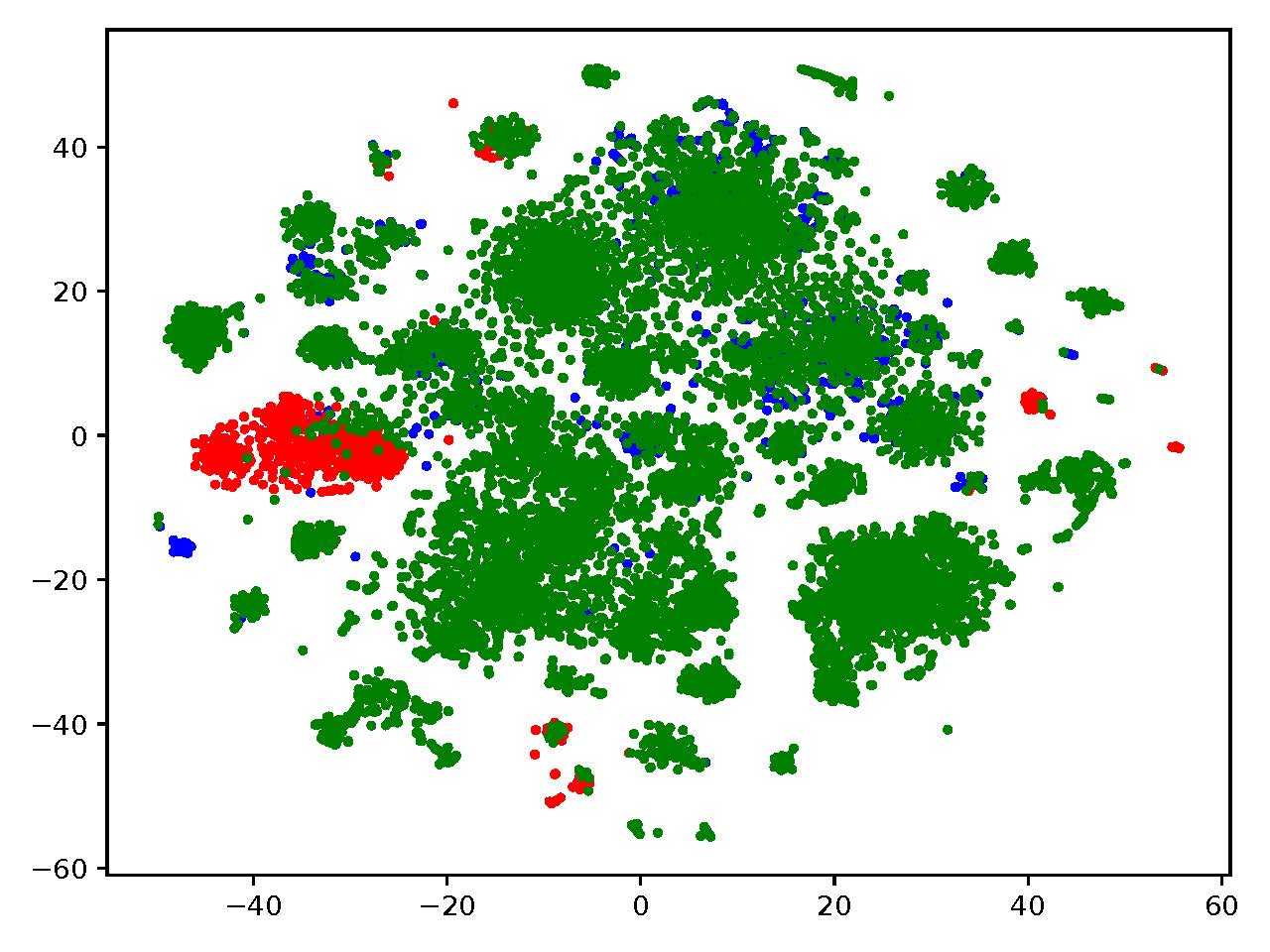}}
\subfigure[FedEC]{
\label{fig:tsne_TransE_c3_FedEC} %
\includegraphics[height = 30 mm]{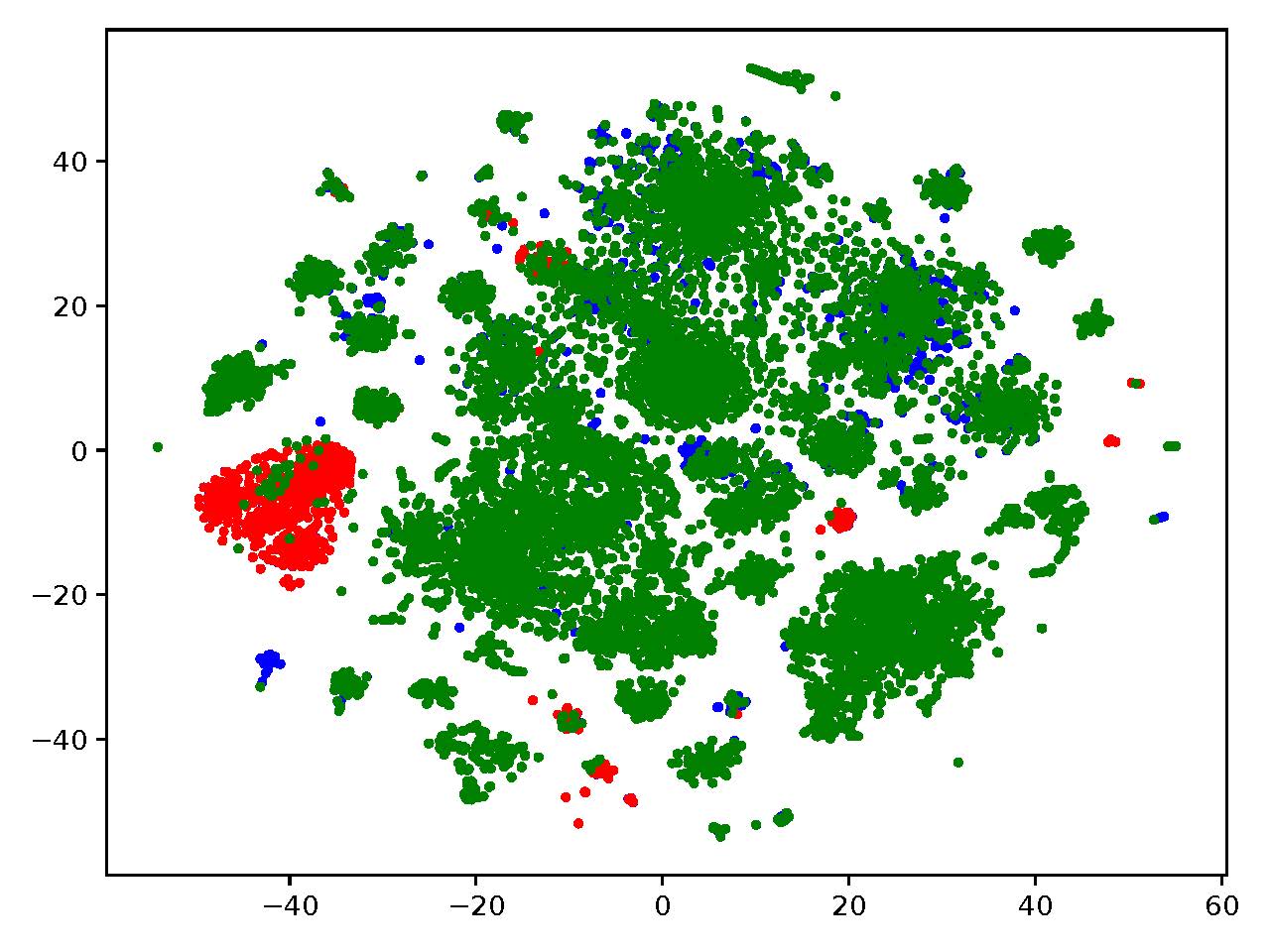}}
\subfigure[\modelname]{
\label{fig:tsne_TransE_c3_FedLU} %
\includegraphics[height = 30 mm]{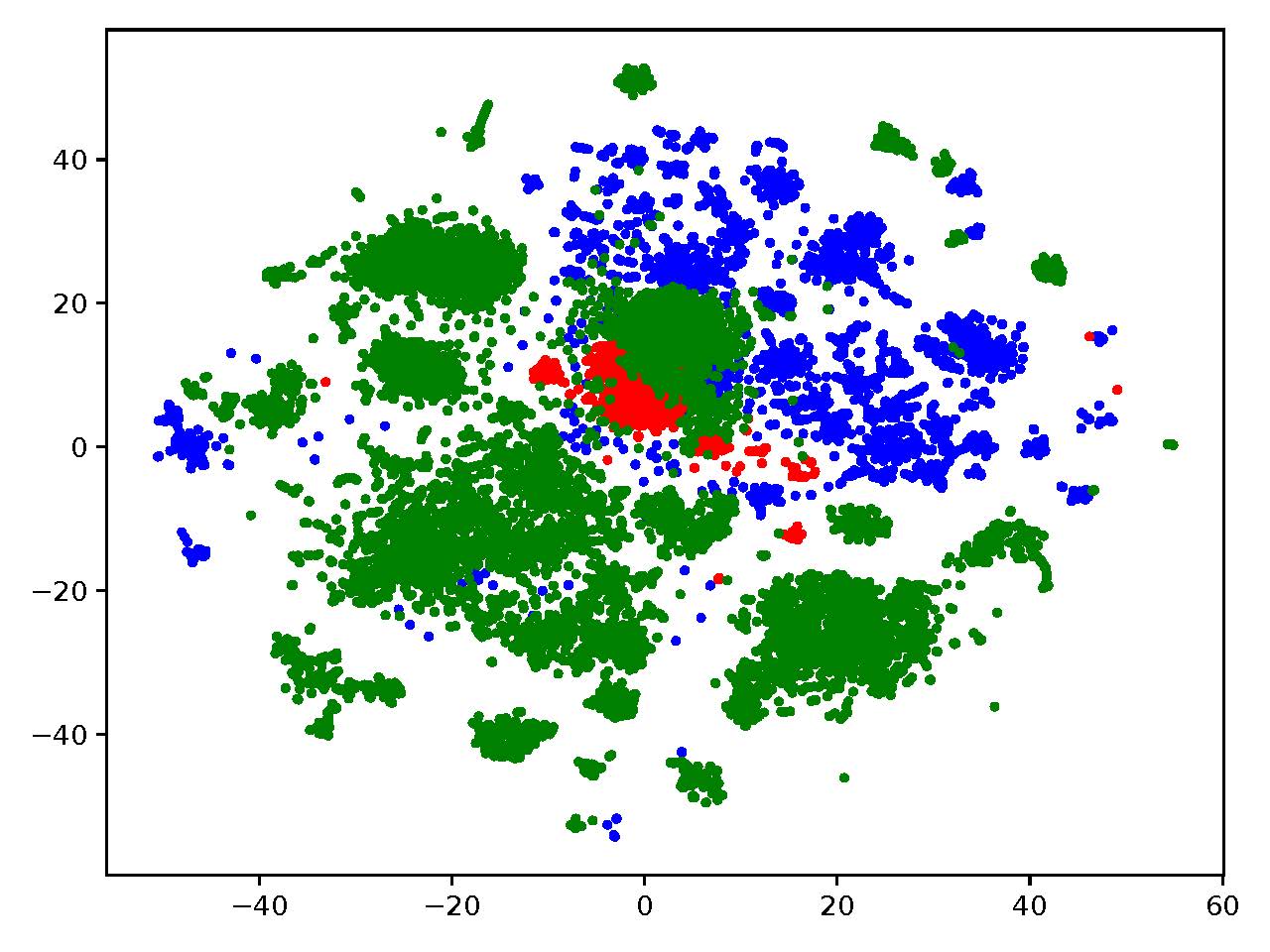}}
\caption{t-SNE plot of entity embeddings on FB15k-237-C3 from each client using TransE, where blue, red and green dots are from Clients 1, 2 and 3, respectively.}
\label{fig:tsne_TransE_c3}
\end{figure}

\smallskip
\noindent\textbf{Entity embedding visualization.}
To find out the reason that \modelname shows advantage against other federated KG embedding methods, we plot the t-SNE \cite{t-SNE} visualization of the entity embeddings on FB15k-237-C3 for each client using TransE in Figure~\ref{fig:tsne_TransE_c3}.
Appendix~\ref{app:entity_embedding_fb237_C5} shows the visualization on FB15k-237-C5.
The triplet proportions in the three clients are approximately 18:5:80.
We observe that FedE, FedProx and FedEC exhibit commonalities in their embeddings: entities on Client 2 (with less triplets) are isolated, while entities on Client 1 and Client 3 are highly overlapping.
This indicates that, for better performance in KG embedding, FedE, FedProx and FedEC tend to aggregate knowledge of the clients with more triplets while ignoring clients with less triplets.
However, \modelname shows a quite different manner.
The entity embeddings between each client intersect with each other, but only have small overlap.
Also, the entity embeddings within each client are more clustered.
This indicates that the entity embeddings generated by \modelname can not only relate and aggregate the knowledge of all clients, but also give consideration to the local optimization.
\modelname achieves better performance than others as the learned embeddings describe different but related sides of corresponding entities.

%==============================
\section{Conclusion}

We propose \modelname, a novel federated KG embedding learning and unlearning framework.
In federated learning, we design mutual knowledge distillation to smoothly exchange knowledge among clients, achieving global convergence and local optimization.
In federated unlearning, we combine retroactive interference and passive decay to enable triplet unlearning and propagation.
We conduct extensive experiments on three newly-constructed datasets of heterogeneity.
The experimental results show that \modelname achieves better accuracy in link prediction. 
It can also forget specific knowledge without significantly hurting the overall performance.
In future work, we plan to study the continual learning and unlearning of federated KG embedding along the life cycle of KGs.
%We also want to leverage pre-trained language models to incorporate literal knowledge for mutual distillation.

\smallskip\noindent\textbf{Acknowledgments.} 
This work is supported by National Natural Science Foundation of China (No. 62272219).

%==============================
%\clearpage
%\balance
\bibliographystyle{ACM-Reference-Format}
\bibliography{reference}

%==============================
\clearpage
\appendix

%==============================
\section{Dataset Construction Algorithm}
\label{app:algo}

Algorithm~\ref{alg:dataset_construction} presents the pseudo-code.
In Lines 3-5, we record the set of relations linked by each entity as $E2R$.
In Lines 6-10, we calculate the co-occurrence matrix $M$ of relations based on $E2R$.
In Line 11, we select a spectral clustering algorithm \cite{SpectralClusteringSurvey} to partition relations.
Specifically, the Laplacian matrix $L$ of $M$ is first computed.
Then, the first $|K|$ eigenvalues and corresponding eigenvectors of $L$ are solved and formed into matrix $S\in\mathbb{R}_{|R|\times|K|}$, each row of which denotes the feature of a relation in a $|K|$-dimensional space.
Next, the k-means clustering algorithm is conducted on $S$ to divide $R$ into $|K|$ clusters.
Finally, the triplets are distributed in Line 12.

\begin{algorithm}[!ht]
\caption{Clustering-based dataset construction}
\label{alg:dataset_construction}
  \KwIn{original KG $G=\{(h,r,t)\}$, client number $|K|$}
  \KwOut{dataset $D=\{G^k\}_{k=1}^{|K|}$, s.t. $G^k\subset G, \bigcup_{k=1}^{|K|} G^{k}=G$}
  $E=\{h\,|\,(h,r,t)\in G\}\cup \{t\,|\,(h,r,t)\in G\}$\;
  $R=\{r\,|\,(h,r,t)\in G\}$\;
  $E2R\gets \mathop{Dict}()$ \tcp*{init mappings from ent. to rel.}
  \lForEach{$e\in E$}{$E2R[e]\gets \mathop{Set}()$}
  \lForEach{$(h,r,t)\in G$}{$E2R[h].\mathop{add}(r)$; $E2R[t].\mathop{add}(r)$}
  $M\gets[0]_{|R|\times|R|}$ \tcp*{init co-occurrence matrix}
  \ForEach{$e\in E$}{
    $R'\gets E2R[e]$\;
    \ForEach{$r_1,r_2\in R',r_1\neq r_2$}{
      $M[r_1,r_2]\gets M[r_1,r_2]+1$\;
    }
  }
  $\{R^1,R^2,\dots,R^{|K|}\}\gets \mathop{SpectralClustering}(R,M,K)$\;
  \lForEach{$k=1,\dots,|K|$}{$G^k\gets\{(h,r,t)\in G\,|\,r\in R^k\}$}
\end{algorithm}

%==============================
\section{Dataset Comparison}
\label{app:dataset}

\begin{figure}[!b]
  \centering
  \includegraphics[width=\linewidth]{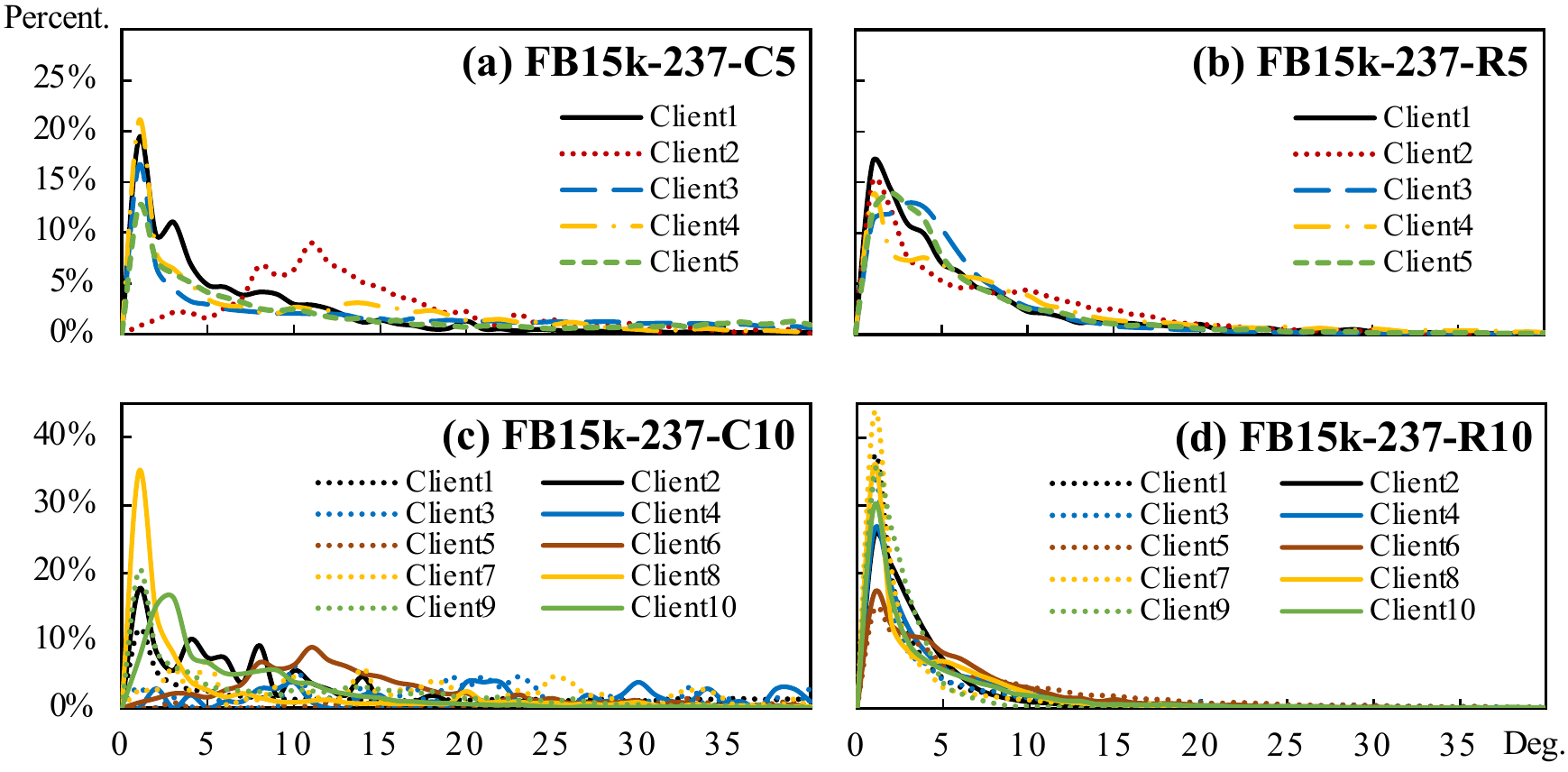}
  \caption{Comparison of degree distributions between FB15k-237-C5/C10 and FB15k-237-R5/R10}
  \label{fig:degree2}
\end{figure}

Figure~\ref{fig:degree2} shows the comparison results of the degree distributions between FB15k-237-C5 and FB15k-237-R5, and between FB15k-237-C10 and FB15k-237-R10. 
We can find that the degree distributions on FB15k-237-C5/C10 vary more significantly.
Such situations are similar to those on FB15k-237-C3 and FB15k-237-R3.
%It again shows that our clustering-based method is more appropriate to construct heterogeneous FL datasets than just randomly partitioning.

%==============================
\section{More Experimental Results}

\begin{figure}[!t]
\centering
\subfigure[FedE]{
\label{fig:tsne_TransE_c5_FedE} %
\includegraphics[height = 30 mm]{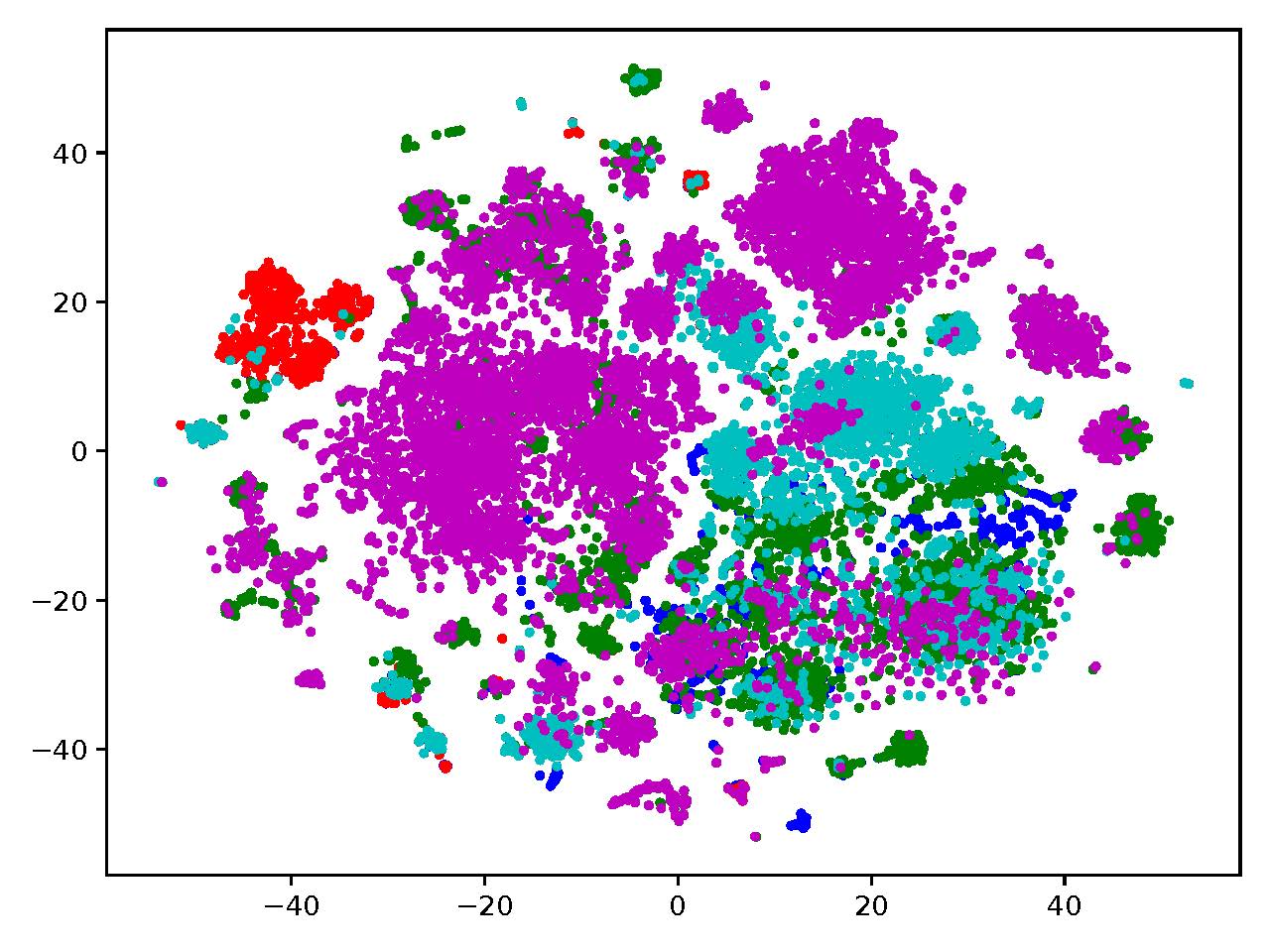}}
\subfigure[FedProx]{
\label{fig:tsne_TransE_c5_FedProx} %
\includegraphics[height = 30 mm]{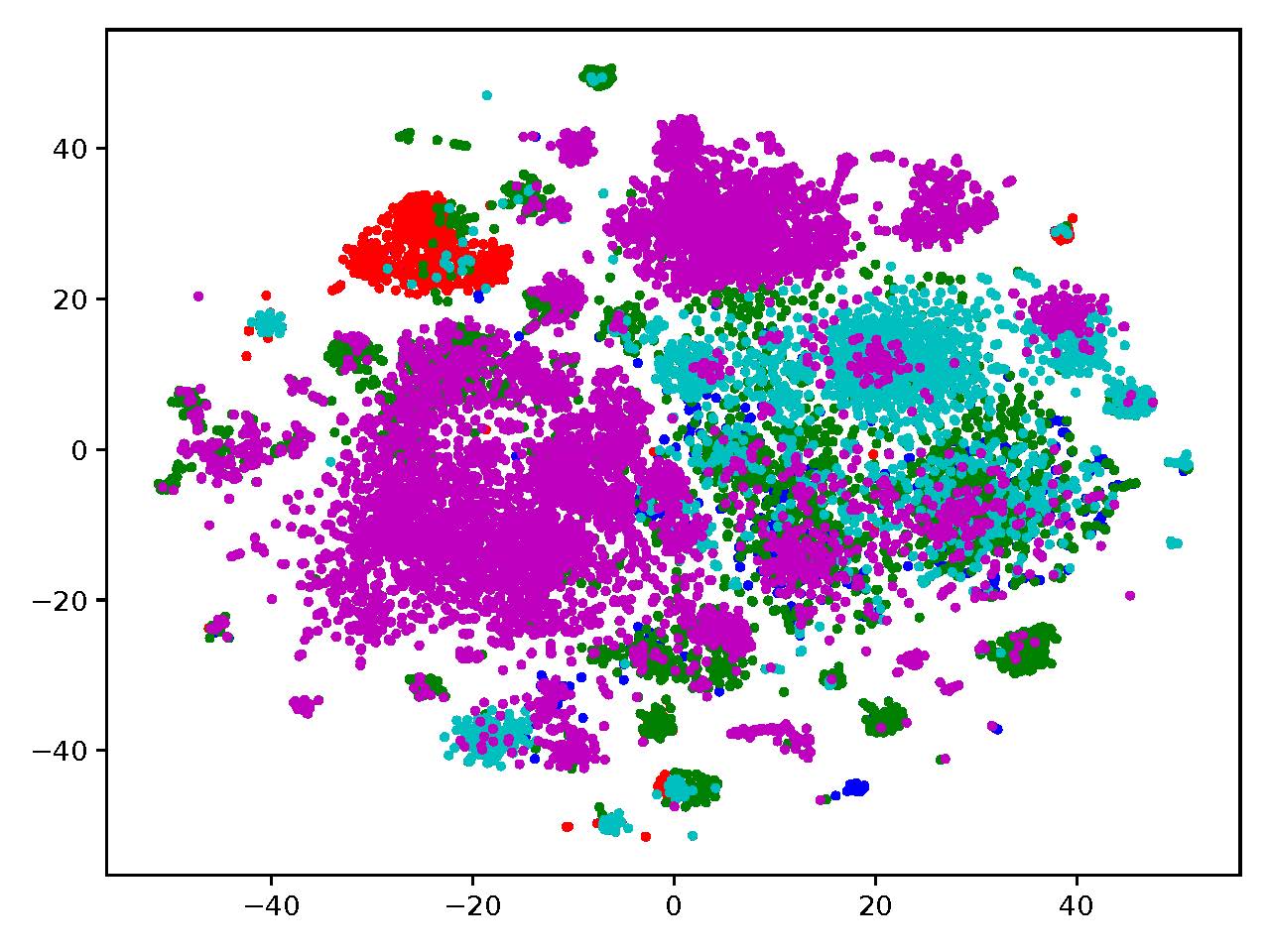}}
\subfigure[FedEC]{
\label{fig:tsne_TransE_c5_FedEC} %
\includegraphics[height = 30 mm]{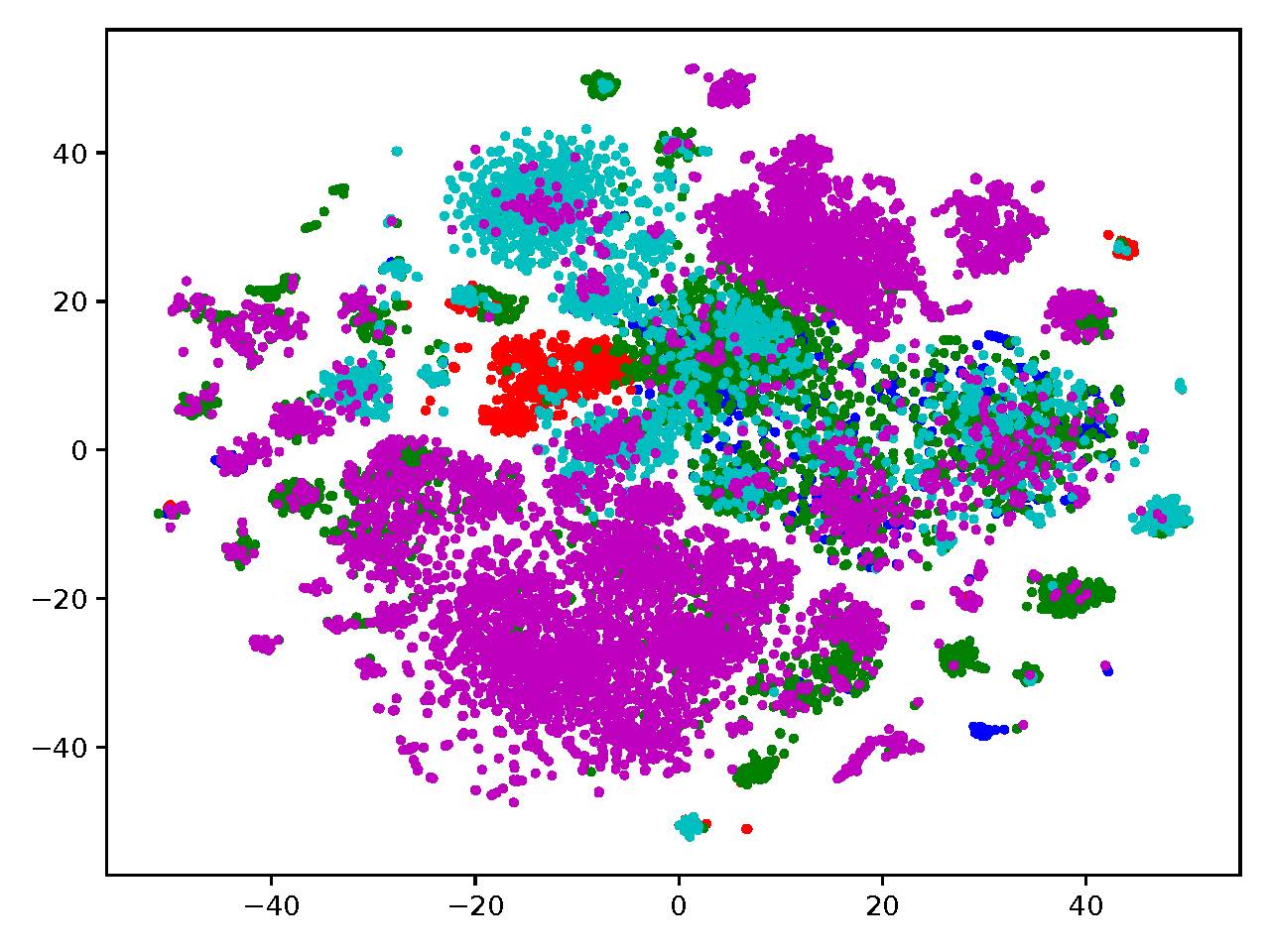}}
\subfigure[\modelname]{
\label{fig:tsne_TransE_c5_FedLU} %
\includegraphics[height = 30 mm]{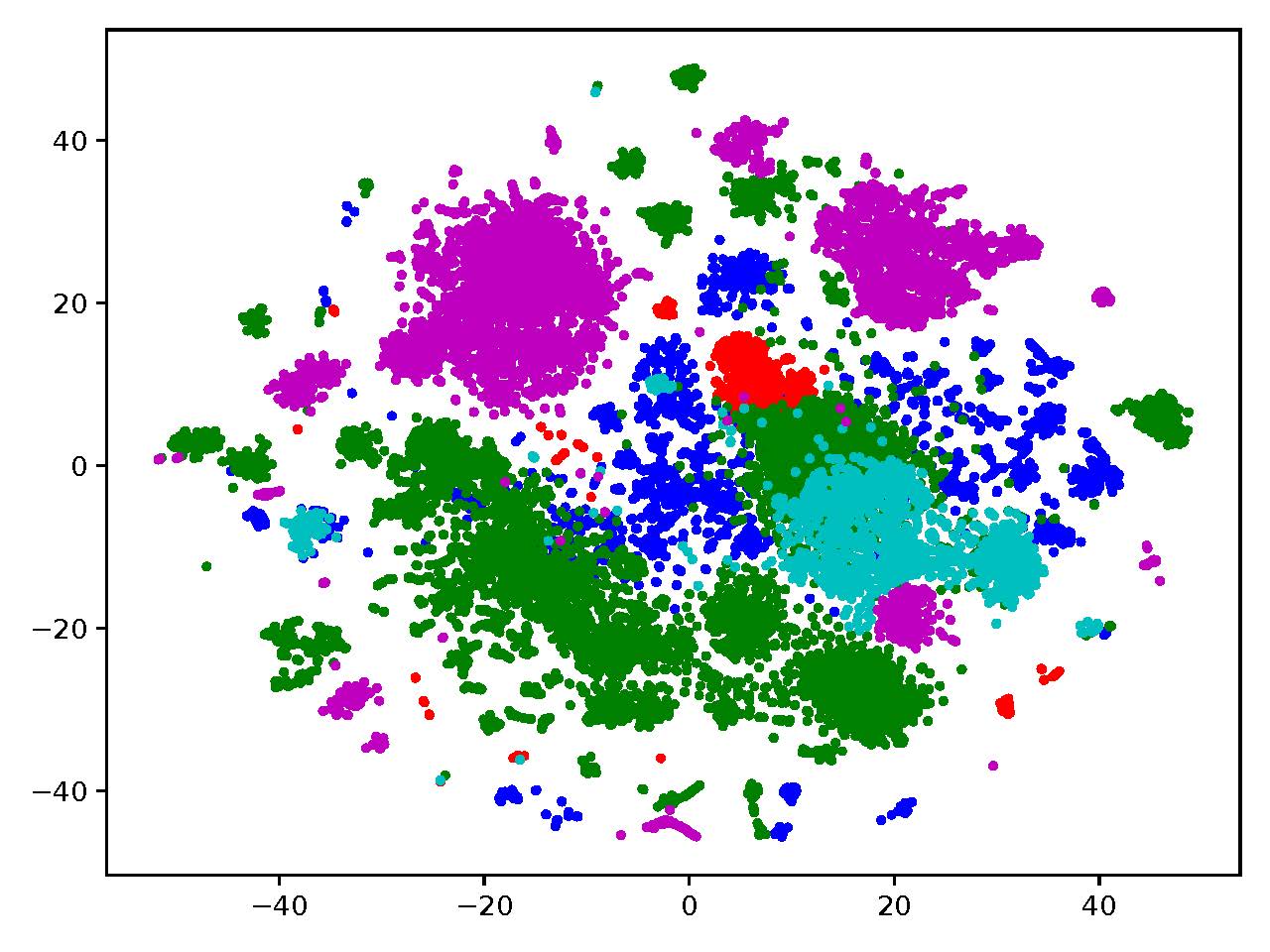}}
\caption{t-SNE plot of entity embeddings on FB15k-237-C5 from each client using TransE, where blue, red, green, cyan and magenta dots are from Clients 1, 2, 3, 4 and 5, respectively.}
\label{fig:tsne_TransE_c5}
\end{figure}

%As a complement to the experimental results reported in the main content, we show more federated KG embedding learning and unlearning results below.

%--------------------
\subsection{Learning Results on FB15k-237-R3/R5/R10}
\label{app:lp_fb237_R}

We also conduct link prediction on FB15k-237-R3, R5 and R10. 
Table~\ref{tab:lp_fb237r} presents the experimental results.
We can observe that both \modelname and existing federated KG embedding methods generally achieve an improvement compared with the independent setting.
\modelname outperforms other competitors stably on all the three datasets, and even beats the centralized setting sometimes.

Compared with the results in Table~\ref{tab:lp_fb237c}, we also find two interesting differences.
First, the gap between the independent setting and the centralized setting is more significant.
This is because FB15k-237-R3/R5/R10 are less heterogeneous, which are easier to obtain noticeable increases with more data. 
%Besides, the improvement from all FL embedding algorithms on FB15k-237-R3/R5/R10 is more considerable. 
%This indicates that our clustering-based datasets are more appropriate to construct difficult datasets for assessing FL algorithms. 
Second, \modelname (local) generally underperforms \modelname (global) in Table~\ref{tab:lp_fb237r}, while it performs better in Table~\ref{tab:lp_fb237c}.
%This is again due to the fact that FB15k-237-C3/C5/C10 are more heterogeneous.
The global embedding stored in \modelname represents the characteristics of the entire data.
They cannot capture the local heterogeneous data on each client well.
FB15k-237-R3/R5/R10 do not encounter such an issue as they are nearly homogeneous.

%--------------------
\subsection{Unlearning Results on FB15k-237-C5/C10}
\label{app:unlearn_fb237_C5C10}

We also conduct federated unlearning on FB15k-237-C5 and FB15k-237-C10.
Tables~\ref{tab:unlearn_fb237c5} and \ref{tab:unlearn_fb237c10} show the average results.
The conclusion is similar to that from Table~\ref{tab:unlearn_fb237c3}.
Compared with the raw results, the results on the forgetting sets decrease rapidly with our unlearning method, while there are only slight declines on the test sets.
We also notice that the re-trained results on the forgetting sets are much higher than our unlearning results.
This not only indicates that simply re-training on the retaining sets cannot achieve unlearning, but also validates that our proposed unlearning method is effective.
%Additionally, we find that the results with local embeddings in \modelname are consistently better than those with global embeddings.
%The reason remains that our constructed datasets based on clustering are more heterogeneous.

%--------------------
\subsection{Entity Embedding Visualization on FB15k-237-C5}
\label{app:entity_embedding_fb237_C5}

Figure~\ref{fig:tsne_TransE_c5} visualizes the entity embeddings for each client on FB15k-237-C5 with TransE.
We also find that the entity embeddings of \modelname on each client are better clustered than other competitors, which demonstrates the superiority of \modelname.

\begin{table*}
\centering
\caption{Average link prediction results on FB15k-237-R3/R5/R10}
{\small
\begin{tabular}{ll|cccc|cccc|cccc}
\toprule 
	\multicolumn{2}{c|}{\multirow{2}{*}{Methods}} & \multicolumn{4}{c|}{FB15k-237-R3} & \multicolumn{4}{c|}{FB15k-237-R5} & \multicolumn{4}{c}{FB15k-237-R10} \\
	\cmidrule(lr){3-6} \cmidrule(lr){7-10} \cmidrule(lr){11-14}
	& & Hits@1 & Hits@3 & Hits@10 & MRR & Hits@1 & Hits@3 & Hits@10 & MRR & Hits@1 & Hits@3 & Hits@10 & MRR \\
\midrule
	\parbox[t]{3mm}{\multirow{7}{*}{\rotatebox[origin=c]{90}{TransE}}} & Independent & 19.86 & 38.15 & 55.12 & 31.96 & 19.02 & 37.29 & 53.65 & 30.98 & 17.02 & 36.40 & 53.50 & 29.61 \\
	& Centralized & 20.17 & 38.56 & 56.01 & 32.44 & 20.34 & 38.90 & 56.60 & 32.74 & 19.48 & 38.43 & 56.86 & 32.21 \\
	& FedE & 20.21 & 38.77 & 56.33 & 32.52 & 20.11 & 38.70 & 56.24 & 32.45 & 19.35 & 38.57 & 56.51 & 32.10 \\
	& FedProx & 20.02 & 38.71 & 55.98 & 32.38 & 19.97 & 38.54 & 55.42 & 32.25 & 18.57 & 37.89 & 55.02 & 31.24 \\
	& FedEC & 19.96 & 38.72 & 56.30 & 32.40 & 20.27 & 39.03 & 56.32 & 32.66 & 19.19 & 38.82 & \underline{56.93} & 32.13 \\
	& \modelname (local) & \underline{20.88} & \underline{39.42} & \underline{56.63} & \underline{33.12} & \underline{20.49} & \underline{39.68} & \underline{56.88} & \underline{33.05} & \underline{19.99} & \underline{39.33} & 56.79 & \underline{32.69} \\
	& \modelname (global) & \textbf{22.06} & \textbf{41.62} & \textbf{58.74} & \textbf{34.75} & \textbf{21.35} & \textbf{41.52} & \textbf{59.06} & \textbf{34.38} & \textbf{21.53} & \textbf{41.98} & \textbf{59.78} & \textbf{34.77} \\
\midrule
	\parbox[t]{3mm}{\multirow{7}{*}{\rotatebox[origin=c]{90}{ComplEx}}} & Independent & 21.51 & 34.06 & 47.51 & 30.31 & 21.43 & 33.71 & 46.67 & 30.04 & 20.68 & 33.61 & 47.08 & 29.64 \\
	& Centralized & \textbf{23.42} & \underline{39.89} & \underline{56.23} & \textbf{34.54} & \textbf{24.42} & \textbf{41.03} & \textbf{57.29} & \textbf{35.53} & \textbf{24.71} & \textbf{41.00} & \textbf{57.83} & \textbf{35.83} \\
	& FedE & 20.29 & 36.31 & 52.69 & 31.22 & 18.57 & 35.18 & 51.78 & 29.83 & 17.64 & 34.25 & \underline{51.45} & 29.06 \\
	& FedProx & 22.18 & 39.65 & 56.09 & 33.78 & 17.02 & 29.55 & 43.42 & 25.89 & 16.78 & 28.72 & 42.74 & 25.47 \\
	& FedEC & 18.07 & 31.08 & 45.15 & 27.19 & 17.51 & 30.51 & 44.53 & 26.62 & 17.54 & 31.78 & 47.95 & 27.75 \\
	& \modelname (local) & \underline{22.79} & 39.83 & 56.22 & 34.15 & \underline{21.91} & 38.41 & 54.14 & 32.94 & \underline{22.02} & \underline{35.88} & 49.78 & \underline{31.52} \\
	& \modelname (global) & 22.64 & \textbf{40.28} & \textbf{56.70} & \underline{34.28} & 21.25 & \underline{38.73} & \underline{55.77} & \underline{32.99} & 18.92 & 33.70 & 50.00 & 29.44 \\
\midrule
	\parbox[t]{3mm}{\multirow{7}{*}{\rotatebox[origin=c]{90}{RotatE}}} & Independent & 23.10 & 39.01 & 54.65 & 33.84 & 21.97 & 38.19 & 53.80 & 32.83 & 21.10 & 37.06 & 52.91 & 31.90 \\
	& Centralized & 25.07 & 44.17 & \underline{61.36} & 37.47 & \textbf{25.28} & \textbf{44.63} & \textbf{62.08} & \textbf{37.89} & 24.24 & 44.66 & \underline{62.44} & 37.39 \\
	& FedE & 20.82 & 38.99 & 56.23 & 32.94 & 24.65 & 44.04 & \underline{61.40} & 37.28 & 23.39 & 43.69 & 61.28 & 36.44 \\
	& FedProx & 25.42 & 43.91 & 61.05 & 37.59 & 24.95 & 44.05 & 60.87 & 37.28 & 22.87 & 42.33 & 59.50 & 35.46 \\
	& FedEC & \underline{25.63} & \underline{44.42} & \underline{61.36} & \underline{37.90} & 25.08 & 44.27 & 61.05 & 37.51 & 23.81 & 43.73 & 61.32 & 36.72 \\
	& \modelname (local) & \textbf{25.82} & \textbf{45.35} & \textbf{61.98} & \textbf{38.36} & \underline{25.20} & 44.54 & 61.14 & \underline{37.82} & \underline{24.27} & \textbf{44.99} & \textbf{62.45} & \textbf{37.52} \\	
	& \modelname (global) & 25.29 & 43.33 & 59.93 & 37.15 & 24.23 & \underline{44.56} & 61.18 & 37.24 & \textbf{24.29} & \underline{44.82} & 62.42 & \underline{37.47} \\
\bottomrule
    \multicolumn{14}{l}{The best and second best scores are marked in \textbf{bold} and with \underline{underline}, respectively.}
\end{tabular}}
\label{tab:lp_fb237r}
\end{table*}

\begin{table*}
\centering
\caption{Average unlearning results on different sets of FB15k-237-C5 with the local and global models}
{\small
\begin{tabular}{l|cccccc|cccccc}
\toprule 
	\multirow{2}{*}{Hits@1} & \multicolumn{2}{c}{Raw local} & \multicolumn{2}{c}{Re-trained local} & \multicolumn{2}{c|}{Unlearned local} & \multicolumn{2}{c}{Raw global} & \multicolumn{2}{c}{Re-trained global} & \multicolumn{2}{c}{Unlearned global}\\
	\cmidrule(lr){2-3} \cmidrule(lr){4-5} \cmidrule(lr){6-7} \cmidrule(lr){8-9} \cmidrule(lr){10-11} \cmidrule(lr){12-13} 
	& Forget$\downarrow$ & Test$\uparrow$ & Forget$\downarrow$ & Test$\uparrow$ & Forget$\downarrow$ & Test$\uparrow$ & Forget$\downarrow$ & Test$\uparrow$ & Forget$\downarrow$ & Test$\uparrow$ & Forget$\downarrow$ & Test$\uparrow$ \\
\midrule
	TransE & 43.24 & 21.89 & 19.83 & 21.08 & 12.72 & 21.03 & 37.92 & 21.73 & 19.32 & 20.03 & 13.08 & 21.12 \\
	ComplEx & 60.37 & 24.05 & 17.87 & 19.01 & \ \ 3.60 & 21.85 & 54.37 & 22.03 & 23.01 & 21.93 & \ \ 2.70 & 21.15 \\
	RotatE & 69.65 & 27.38 & 25.10 & 26.82 & 17.61 & 26.86 & 64.27 & 27.07 & 25.66 & 25.83 & 17.57 & 26.68 \\
\bottomrule
\toprule 
	\multirow{2}{*}{MRR} & \multicolumn{2}{c}{Raw local} & \multicolumn{2}{c}{Re-trained local} & \multicolumn{2}{c|}{Unlearned local} & \multicolumn{2}{c}{Raw global} & \multicolumn{2}{c}{Re-trained global} & \multicolumn{2}{c}{Unlearned global}\\
	\cmidrule(lr){2-3} \cmidrule(lr){4-5} \cmidrule(lr){6-7} \cmidrule(lr){8-9} \cmidrule(lr){10-11} \cmidrule(lr){12-13} 
	& Forget$\downarrow$ & Test$\uparrow$ & Forget$\downarrow$ & Test$\uparrow$ & Forget$\downarrow$ & Test$\uparrow$ & Forget$\downarrow$ & Test$\uparrow$ & Forget$\downarrow$ & Test$\uparrow$ & Forget$\downarrow$ & Test$\uparrow$ \\
\midrule
	TransE & 56.58 & 34.32 & 32.21 & 33.34 & 22.71 & 33.40 & 51.98 & 34.30 & 31.53 & 32.26 & 23.54 & 33.58  \\
	ComplEx & 71.14 & 35.66 & 28.64 & 29.72 & \ \ 9.35 & 33.47 & 65.68 & 34.07 & 35.44 & 34.33 & \ \ 8.21 & 32.85 \\
	RotatE & 78.69 & 38.93 & 36.77 & 38.39 & 29.80 & 38.55 & 74.41 & 39.11 & 37.61 & 38.04 & 29.73 & 38.52 \\
\bottomrule
\end{tabular}
}
\label{tab:unlearn_fb237c5}
\end{table*}

\begin{table*}
\centering
\caption{Average unlearning results on different sets of FB15k-237-C10 with the local and global models}
{\small
\begin{tabular}{l|cccccc|cccccc}
\toprule 
	\multirow{2}{*}{Hits@1} & \multicolumn{2}{c}{Raw local} & \multicolumn{2}{c}{Re-trained local} & \multicolumn{2}{c|}{Unlearned local} & \multicolumn{2}{c}{Raw global} & \multicolumn{2}{c}{Re-trained global} & \multicolumn{2}{c}{Unlearned global}\\
	\cmidrule(lr){2-3} \cmidrule(lr){4-5} \cmidrule(lr){6-7} \cmidrule(lr){8-9} \cmidrule(lr){10-11} \cmidrule(lr){12-13} 
	& Forget$\downarrow$ & Test$\uparrow$ & Forget$\downarrow$ & Test$\uparrow$ & Forget$\downarrow$ & Test$\uparrow$ & Forget$\downarrow$ & Test$\uparrow$ & Forget$\downarrow$ & Test$\uparrow$ & Forget$\downarrow$ & Test$\uparrow$ \\
\midrule
	TransE & 39.81 & 21.05 & 18.76 & 18.59 & 13.61 & 20.33 & 35.65 & 20.58 & 20.65 & 20.94 & 14.42 & 20.12 \\
	ComplEx & 57.54 & 25.14 & 15.84 & 16.62 & 10.29 & 24.12 & 49.12 & 22.07 & 23.28 & 23.66 & \ \ 6.65 & 21.65 \\
	RotatE & 64.82 & 26.01 & 24.95 & 24.47 & 16.37 & 25.18 & 60.62 & 25.71 & 24.17 & 24.12 & 18.28 & 25.14 \\
\bottomrule
\toprule 
	\multirow{2}{*}{MRR} & \multicolumn{2}{c}{Raw local} & \multicolumn{2}{c}{Re-trained local} & \multicolumn{2}{c|}{Unlearned local} & \multicolumn{2}{c}{Raw global} & \multicolumn{2}{c}{Re-trained global} & \multicolumn{2}{c}{Unlearned global}\\
	\cmidrule(lr){2-3} \cmidrule(lr){4-5} \cmidrule(lr){6-7} \cmidrule(lr){8-9} \cmidrule(lr){10-11} \cmidrule(lr){12-13} 
	& Forget$\downarrow$ & Test$\uparrow$ & Forget$\downarrow$ & Test$\uparrow$ & Forget$\downarrow$ & Test$\uparrow$ & Forget$\downarrow$ & Test$\uparrow$ & Forget$\downarrow$ & Test$\uparrow$ & Forget$\downarrow$ & Test$\uparrow$ \\
\midrule
	TransE & 53.90 & 33.44 & 30.69 & 30.78 & 24.29 & 32.68 & 49.60 & 32.95 & 33.85 & 34.18 & 25.33 & 32.58 \\
	ComplEx & 67.76 & 36.33 & 27.02 & 27.77 & 18.71 & 35.22 & 59.91 & 33.45 & 35.57 & 36.03 & 13.29 & 32.66 \\
	RotatE & 75.16 & 38.22 & 36.87 & 37.36 & 28.20 & 37.37 & 71.27 & 38.00 & 35.50 & 35.50 & 30.69 & 37.36 \\
\bottomrule
\end{tabular}
}
\label{tab:unlearn_fb237c10}
\end{table*}

\end{document}